\def\year{2020}\relax
\documentclass[letterpaper]{article} 
\usepackage{aaai20}  
\usepackage{pifont}
\usepackage{times}  
\usepackage{helvet} 
\usepackage{courier}  
\usepackage[hyphens]{url}  
\usepackage{subcaption}
\usepackage{graphicx} 
\usepackage{amsfonts}
\usepackage{amssymb}
\usepackage{enumitem}
\usepackage{algorithm}
\usepackage{algorithmicx}
\usepackage{algpseudocode}
\usepackage{amsmath,bm}
\usepackage{graphicx}  
\newcommand{\infsetstyle}[1]{{\mathbb{#1}}}
\newcommand{\finsetstyle}[1]{{\mathbb{#1}}}
\newcommand{\diststyle}[1]{{\mathcal{#1}}}
\newcommand{\gp}[0]{\diststyle{GP}}
\newcommand{\tsp}[0]{{\rm T}}

\newcommand{\latvec}[1]{{\mbox{\boldmath $#1$}}}

\newcommand{\normdist}[0]{\mathcal{N}}

\DeclareMathOperator\argmax{argmax}

\DeclareMathOperator*\maxx{max}
\DeclareMathOperator*\minn{min}
\DeclareMathOperator*\argmaxx{argmax}
\newtheorem{def_preforder}{Definition}
\newtheorem{th_geomofS}{Theorem}
\newtheorem{cor_prop}{Proposition}
\newtheorem{cor_sup}{Corollary}
\newenvironment{proof}{\paragraph{Proof:}}{\hfill$\square$}
\title{Cost-aware Multi-objective Bayesian optimisation}
\author{{Majid Abdolshah, Alistair Shilton, Santu Rana, Sunil Gupta, Svetha Venkatesh}\\ 
\textsuperscript{ }Applied Artificial Intelligence Institute ($\mathrm{A^2I^2}$), Deakin University\\ 
{\{majid, alistair.shilton, santu.rana, sunil.gupta, svetha.venkatesh\}@deakin.edu.au} 
}
 
\begin{document}

\maketitle

\begin{abstract}
The notion of ``expense'' in Bayesian optimisation generally refers to the uniformly expensive cost of function evaluations over the whole search space. However, in some scenarios, the cost of evaluation for black-box objective functions is non-uniform since different inputs from search space may incur different costs for function evaluations.
We introduce a cost-aware multi-objective Bayesian optimisation with non-uniform evaluation cost over objective functions by defining cost-aware constraints over the search space. The cost-aware constraints are a sorted tuple of indexes that demonstrate the \textit{ordering} of dimensions of the search space based on the user's prior knowledge about their cost of usage. We formulate a new multi-objective Bayesian optimisation acquisition function with detailed analysis of the convergence that incorporates this cost-aware constraints while optimising the objective functions. We demonstrate our algorithm based on synthetic and real-world problems in hyperparameter tuning of neural networks and random forests. 
\end{abstract}
\section{Introduction} 
\begin{figure}[t]
\centering
\includegraphics[width=0.41\textwidth]{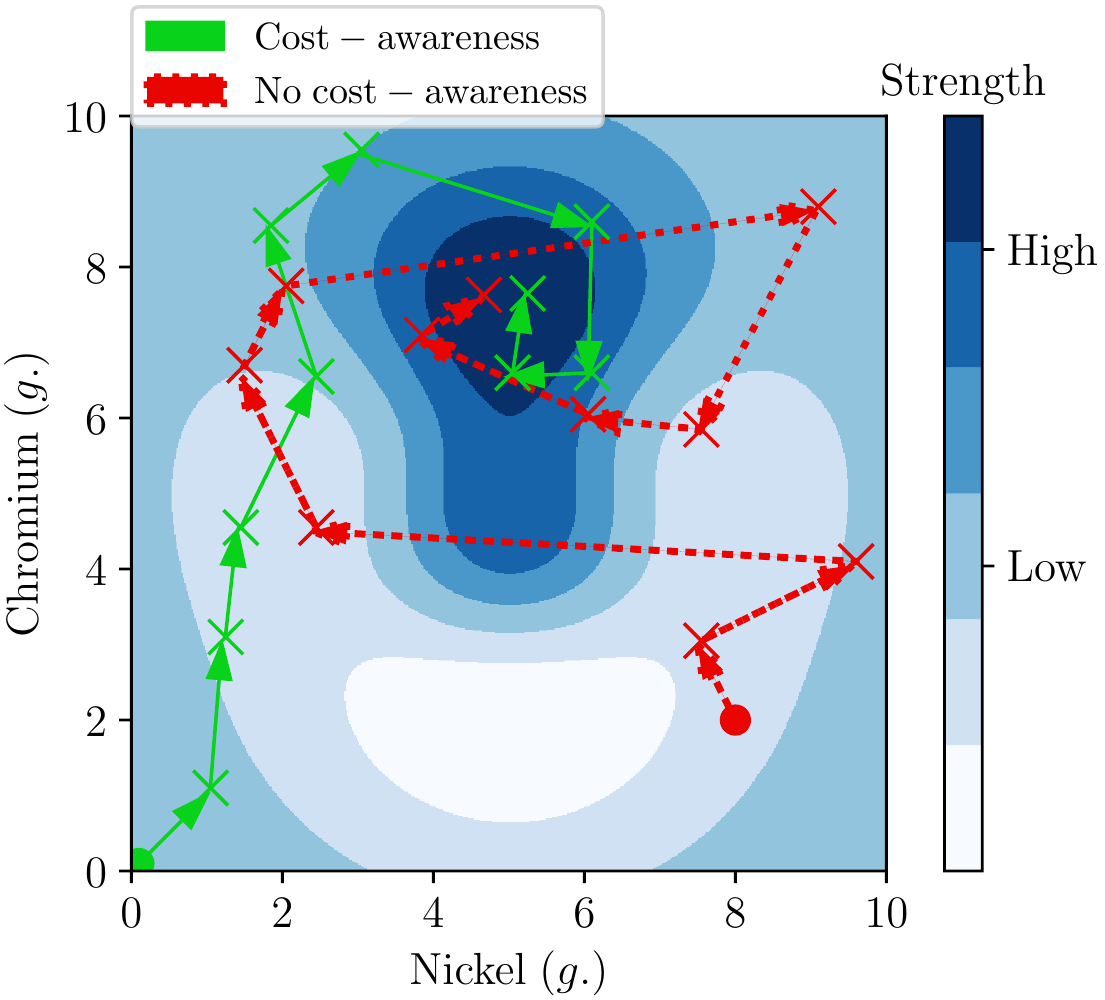}
\caption{Searching for optimal alloy composition to maximise strength, given component Nickel is more expensive than Chromium. Optimisation paths with and without cost-awareness are shown. ``$\bullet$'' are initial observation and ``$\times$'' indicates the remaining points.} \label{fig:motiv}
\end{figure}
\indent{}\par Bayesian optimisation is a well-known algorithm for optimising black-box and expensive to evaluate functions. Multi-objective Bayesian optimisation is a generalised form with  multiple conflicting objectives to be optimised simultaneously \cite{khan2002multi}. 
In such scenarios, the optimiser is seeking for a set of Pareto optimal outcomes often called the Pareto front. Finding Pareto front is an expensive procedure. The notion of \textit{expense} in single/multi-objective Bayesian optimisation refers to  uniform  cost of function evaluations. In some real-world situations, however, the cost of function evaluation may not be uniform because of differential costs across inputs in the search space.  

As an example, consider designing steel using Nickel and Chromium as two key minority ($<20\%$) ingredients in the mix where the rest is Iron. We may consider this differential component cost: ``Nickel is very expensive compared to Chromium'' 
and we would like to find steel with 
 high yield strength.
A plain Bayesian optimisation will progress without  awareness of the costs and may sample anywhere in the search space. The judicious approach is to start with small amounts of the expensive component (Nickel) initially and gradually increase it if the user is not satisfied with solutions obtained with higher quantities of the cheaper component (Chromium). The optimisation path in a single-objective case  with and without awareness of costs is shown in Figure 1. Both paths found the optimum, but the cost-aware path uses a much lower amount of Nickel ($32.1\ g$ ) compared to the vanilla option with no cost-awareness ($54.1\ g$).

\indent{}\par
Such non-uniform cost of function evaluation has not been investigated in the context of Bayesian optimisation. 
The closest related works investigate budget constraints leading to the need to find solutions within specified number of iterations  \cite{hoffman2014correlation,lam2016bayesian,li2019bayesian}. This problem is quite different and the solutions do not translate. There are studies based on evolutionary methods related to cost-effective search space optimisation but these are out of the context in Bayesian optimisation and our proposed problem since their precise definition of cost vary for each case of study such as cache allocation and assignment review \cite{li2017multi,wang2017cost}.
\indent{}\par
To formalise this notion of non-uniform evaluation cost of objective functions, we define a new constraint on the search space (independent of the objective space) based on user's prior knowledge . This knowledge is of the form: \textit{``dimension $i$ of the search space is more expensive than dimension $j$''}. We term these as  \textit {cost-aware preferences} over the input domain. Our goal is to formulate  a solution to incorporate non-uniform cost of functions evaluation  through a Cost-aware Multi-objective Bayesian Optimisation (CA-MOBO) acquisition function. Our motivation follows:

\begin{enumerate}
\item Initial optimiser suggestions  should avoid expensive inputs in search space since a desirable solution may be possible with a cheaper combination of  inputs - as an example, Figure \ref{fig:motiv} shows a cost-aware optimiser starting from a cheap combination.
\item As optimisation progresses, the influence of cost-aware constraints must diminish to ensure the optimiser is able to eventually find all Pareto front solutions.
\end{enumerate}
Our proposed acquisition function has two criteria: Chebyshev scalarisation for objective functions to ensure the solutions satisfy Pareto optimality, and a cost function as a component of the acquisition function that incorporates the user's prior knowledge of the search space.
 Our algorithm (CA-MOBO) favors  solutions that comply with the cost-aware constraints whilst finding  ``good'' quality solutions in the optimisation process. 
\par
Our main contributions are: 
\begin{itemize}[label=\textbullet]
\item Construction of a new Bayesian optimisation framework to incorporate non-uniform cost of function evaluations;
\item Design of a new acquisition function that scales linearly with the number of objective functions. This is significant as comparing to hypervolume based methods, this algorithm is able to tackle many-objective problems; 
\item Definition of regret for this problem, and theoretical proof that it is upper bounded; and,
\item Demonstration of the method on real and synthetic experiments.
\end{itemize}
\section{Notation}
$\infsetstyle{Z}_+ = \{ 1, 2, \ldots \}$, $\infsetstyle{Z}_n = \{ 0, 1, \ldots, n-1 \}$, and $\infsetstyle{Z}^+_n = \{ 1, 2, \ldots, n\}$. $\mathbb{X}$ is the search space, $\mathbb{D}$ is the set of observations and $\infsetstyle{R}$ is the set of real numbers. $|\infsetstyle{X}|$   is the cardinality of the set ${\infsetstyle{X}}$. Tuples (ordered sets) are denoted $\mathcal{A}, \mathcal{B}, \mathcal{C}, \ldots$. Column vectors are bold lower case ${\bf a}, {\bf b}, {\bf c}, \ldots$ and  matrices bold upper case ${\bf A}, {\bf B}, {\bf C}, \ldots$. Element $i$ of vector ${\bf a}$ is $a_i$, and element $i,j$ of matrix ${\bf A}$ is $A_{i,j}$ (all indexed $i,j = 1,2,\ldots$).
\section{Background}
\subsection{Gaussian Processes} \label{sec:gp_intro}
We briefly review Gaussian process (GP) \cite{Ras2}. Given that $\infsetstyle{X} \subset \infsetstyle{R}^N$ is compact,  a Gaussian process  $\gp ( \mu, K)$ is a distribution on the function space $f : 
\infsetstyle{X} \to \infsetstyle{R}$ defined by mean $\mu : \infsetstyle{X} \to 
\infsetstyle{R}$ (assumed zero without loss of generality) and kernel 
(covariance) $K : \infsetstyle{X} \times \infsetstyle{X} \to \infsetstyle{R}$.  
If $f ({\bf x}) \sim \gp ( 0, K ({\bf x}, {\bf x}') )$ then the posterior of 
$f$ given 
$\finsetstyle{D} = \{ ({\bf x}_{j}, y_{j}) \in \infsetstyle{R}^N \times 
\infsetstyle{R} | y_{j} = f ({\bf x}_{j}) + \epsilon, \epsilon \sim \normdist (0, 
\sigma_{noise}^2), j \in \infsetstyle{Z}^+_N \}$ is $f ({\bf x}) | \finsetstyle{D} \sim 
\normdist ( \mu ({\bf x}), {\sigma^2} ({\bf 
x}, {\bf x}'))$, where:
\begin{equation}
 \!\!\!\begin{array}{rl}
 \mu \left( {\bf x} \right) 
   &\!\!\!= {\bf k}^{\tsp} \left( {\bf x} \right) \left( {\bf K} + \sigma_{noise}^2 {\bf I} \right)^{-1} {\bf y} \\
 \sigma^2 \left( {\bf x}, {\bf x}' \right) 
 &\!\!\!= K \left( {\bf x}, {\bf x}' \right) - {\bf k}^{\tsp} \left( {\bf x} \right) \left( {\bf K} + \sigma_{noise}^2 {\bf I} \right)^{-1} {\bf k} \left( {\bf x}' \right) 
 \end{array}\!\!\!\!\!\!\!\!
 \label{eq:gp}
\end{equation}
and  ${\bf y}, {\bf k} ({\bf x}) \in \infsetstyle{R}^{|{\infsetstyle{D}}|}$, ${\bf K} \in 
\infsetstyle{R}^{{|\infsetstyle{D}|} \times {|{ \infsetstyle{D}}|}}$, $k ({\bf x})_j = K ({\bf x}, {\bf x}_{j})$, $K_{jk} = K ({\bf x}_{j}, {\bf x}_{k})$. Typically in Bayesian optimisation, the black-box, expensive to evaluate objective function is modeled as a draw from a Gaussian process.

\subsection{Multi-Objective Optimisation}
Multi-objective Optimisation (MOO) problem is defined as:
\begin{equation}
 \begin{array}{l}
  \mathop{\argmax}\limits_{{\bf x} \in \infsetstyle{X}} {\bf f} \left( {\bf x} \right) \\
 \end{array}
 \label{eq:moo_problem}
\end{equation}
where the components of ${\bf f} : \infsetstyle{X} \subset \infsetstyle{R}^N 
\to \infsetstyle{Y} \subset \infsetstyle{R}^M$ represent the $M$ distinct 
objectives $f_i : \infsetstyle{X} \to \infsetstyle{R}$.  $\infsetstyle{X}$ and 
$\infsetstyle{Y}$ are called design space and objective space, respectively. These objectives must be optimised simultaneously.
A Pareto-optimal solution is a point ${\bf x}^\star \in \infsetstyle{X}$ for 
which it is not possible to find another solution ${\bf x} \in \infsetstyle{X}$ 
such that $f_i ({\bf x}) > f_i ({\bf x}^\star)$ for all objectives $f_i,\ \forall i \in \infsetstyle{Z}^+_M$.  The set of all Pareto optimal solutions is called the Pareto 
set \cite{deb2001multi}:
\begin{equation}
 \begin{array}{l}
  \infsetstyle{X}^\star = \left\{ \left. {\bf x}^\star \in \infsetstyle{X} \right| \nexists {\bf x} \in \infsetstyle{X} : {\bf f} \left( {\bf x} \right) \succ {\bf f} \left( {\bf x}^\star \right) \right\}
 \end{array}
 \label{eq:pareto_set}
\end{equation}
where ${\bf y} \succ {\bf y}'$ (${\bf y}$ dominates ${\bf y}'$) means ${\bf y} 
\ne {\bf y}'$, $y_i \geq y_i'$ $\forall i$, and ${\bf y} \succeq {\bf y}'$ 
means ${\bf y} \succ {\bf y}'$ or ${\bf y} = {\bf y}'$.
Given observations $\finsetstyle{D} = \{ ({\bf x}_{j}, {\bf y}_{j}) \in 
\infsetstyle{R}^N \times \infsetstyle{R}^M | {\bf y}_{j} = {\bf f} ({\bf x}_{j}) + 
\latvec{\epsilon}, \epsilon_i \sim \normdist (0, \sigma_i^2 )\}$ of ${\bf f}$ the dominant set:
\begin{equation}
 \begin{array}{l}
  \finsetstyle{D}^\ast = \left\{ \left. \left( {\bf x}^\ast, {\bf y}^\ast \right) \in \finsetstyle{D} \right| \nexists \left( {\bf x}, {\bf y} \right) \in \finsetstyle{D} : {\bf y} \succeq {\bf y}^\ast \right\}
 \end{array}
 \label{eq:domset}
\end{equation}
is the most optimal subset of $\finsetstyle{D}$ (in the Pareto sense).
\indent{}\par
\subsection{Multi-objective Bayesian Optimisation}\label{sec:MOBO}
Multi-objective Bayesian Optimisation (MOBO) is an iterative optimisation algorithm designed to simultaneously optimise  objective functions that are black-box and expensive to evaluate. At every iteration a sample is selected by maximising a cheap acquisition function $\alpha_t:\ \mathbb{X}\rightarrow \mathbb{R}$ constructed based on a model of $f$, given the previous observations $\mathbb{D}$ \cite{brochu2010tutorial,khan2002multi}. MOBO  aims to obtain the Pareto front in least number of function evaluations.
\indent{}\par
Generally, in MOBO, three approaches are considered to obtain the Pareto front. (a) Predictive Entropy Search \cite{hernandez2016predictive,garrido2019predictive} finds the most expected solution to reduce the entropy of posterior estimate of the Pareto set. (b) Dominated  Hypervolume Improvement ($S$-metric) \cite{Zit1,emmerich2011hypervolume,abdolshah2018expected} maximises the dominated hypervolume (the volume of points in functional space above the Pareto front, with respect to a given reference point ${\bf z} \in \infsetstyle{R}^M$). 
(c) Scalarisation \cite{boyd2004convex} is a standard technique for finding Pareto optimal points by transforming a MOO problem into a single objective optimisation problem $S: \mathbb{R}^M \rightarrow \mathbb{R}$.
\indent{}\par
Many different types of scalarisation functions have been studied in multi-objective optimisation problems \cite{miettinen2002scalarizing,chugh2019scalarizing}. The most simple form of scalarised functions for a multi-objective optimisation problem is weighted sum of the objective functions.
The weighted sum combines different objectives linearly and has been widely used \cite{miettinen2012nonlinear}. However,  it fails for non-convex regions of the Pareto front \cite{emmerich2018tutorial}. Chebyshev scalarisation \cite{borwein1993super} was introduced to overcome this limitation of weighted sum scalarisation function. 
Chebyshev scalarisation function is defined as:
\begin{equation}
S_{{\bm \theta}}(f({\bf x})) = \minn_{m=1}^{M} \theta_m (f({\bf x})_m - R_m) 
\label{eq:cheb}
\end{equation}
where $\mathbf R = [R_1,\ldots,R_M]$ is a preferred reference point and $\bm \theta$ is the weight vector. $\bm \theta$ can be selected randomly or sampled based on a distribution such as ${U}_{\bm \theta}$.
\begin{cor_prop}
For a given set of mutually non-dominated solutions (e.g., a Pareto front) in objective space $\mathbb{R}^M$, for every non-dominated point such as $\mathbf x^\prime$ there exists a set of weight vectors for a Chebyshev scalarisation, that makes this point a maximum of a Chebyshev scalarisation problem provided that the reference point $\mathbf R$ is properly selected.
\label{pro:1}
\end{cor_prop}
As explained and proved in \cite{emmerich2018tutorial}, Proposition \ref{pro:1} ensures   that   by modification of the weight vectors, all points of the Pareto front are also a solution of (\ref{eq:cheb}). As a result, in Chebyshev scalarising function, there exist theoretical results stating that the solutions of (\ref{eq:cheb}) will be at least weakly Pareto optimal for any weighting vector  \cite{emmerich2018tutorial}. 
\indent{}\par
In this study we are defining a new cost-aware acquisition function for multi-objective Bayesian optimisation based on the Chebyshev scalarisation function. 
We will formulate the new algorithm in the next section.
\section{Problem Formulation}
Our proposed acquisition function is based on two criteria:   scalarised Gaussian Process Upper Confidence Bound (GP-UCB), and a cost function that operates as an independent cost-aware agent. We first start with scalarised GP-UCB.
\subsection{Scalarised GP-UCB}
GP-UCB  \cite{srinivas2009gaussian} defines an upper confidence bound as $q_t(\mathbf x) = \mu_{t-1}(\mathbf x) + \sqrt{\beta}_{t}\sigma_{t-1}(\mathbf x)$ where $\mu_{t-1}(\mathbf x)$ and  $\sigma_{t-1}(\mathbf x)$ is defined as in (\ref{eq:gp}) and ${\beta}_{t}$ is a trade-off parameter that grows with $O(log(t))$. In a single-objective BO framework, GP-UCB at each time step $t$, selects the point that maximises  $q_t(\mathbf x)$, i.e. $\mathbf x_t  = \argmaxx_{{\bf x} \in \mathbb{X}}^{ } \ \ q_t(\mathbf x)$.
Recently, an scalarised modification of GP-UCB has been introduced \cite{Kand2018}.
The core idea of scalarised GP-UCB follows from \cite{roijers2013survey,zintgraf2015quality} and the assumption of $S_{{\bm \theta}}(f({\bf x}))$ monotonically  increasing  in  all  coordinates. Given the mentioned assumptions and Proposition \ref{pro:1}, optimising $S_{{\bm \theta}}(f({\bf x}))$ as a maximisation problem returns a single optimal point lying on the Pareto front:
\begin{equation}
{\bf x}^*_{\bm \theta} = \argmaxx_{{\bf x} \in \mathbb{X}}^{ } \ \ S_{{\bm \theta}}(f({\bf x}))
\label{eq:cheb2}
\end{equation}
Generally, MOBO models the objective functions  $f_i,\ \forall i \in \infsetstyle{Z}^+_M$ by sampling from $M$ independent GP posteriors $\mathcal{GP}_{1...M}(\mu^t(\mathbf  x),\sigma^t(\mathbf  x))$. Inspired by \cite{srinivas2009gaussian} and MOBO modeling of objective functions, we can define the scalarised GP-UCB as:
\begin{align}
Q({\bf x},{\bm \theta_t}) = S_{{\bm \theta_t}}(\mu^{t-1}({\bf x}) + \sqrt{\beta_{t}}\sigma^{t-1} ({\bf x}))= ... \nonumber\\ 
 = \minn_{m=1}^{M} \theta_m \Big(\big(\mu_m^{t-1}({\bf x}) + \sqrt{\beta_{t}}\sigma_m^{t-1} ({\bf x})\big) - R_m\Big) 
\label{eq:Q}
\end{align}
$\mu^{t-1}({\bf x})$ and $\sigma^{t-1} ({\bf x})$  are $M$ dimensional vectors denoting the posterior means and variances at $\mathbf x$ for the $M$ objectives respectively at iteration $t-1$. Having defined scalarised GP-UCB, we now introduce CA-MOBO acquisition function.
\subsection{CA-MOBO Acquisition Function}
In our proposed problem, 
for a sample input vector 
as ${\bf x} \in \mathbb{X}$, $x_i$ and $x_j,\  i\neq j$ may incur different costs for the optimiser based on the cost-aware constraints defined by user. In order to calculate the incurred cost of objective function evaluation for input $\mathbf x$, we first formulate cost-aware constraints.
\begin{def_preforder}[Cost-Aware Constraints]
Let $\mathcal{I}=(i_1, i_2,...,i_k)|\{i_1, i_2,...,i_k\} \subset \mathbb{Z}^{+}_N, i_k \neq i_{{k}^\prime} \forall k\neq{k^\prime}$ be a cost-aware constraint over $k$ dimensions of the search space $(1\leq k\leq N)$. Then selecting $x_{(i_{j})}$ value as a input from dimension $i_j$ of the search space is more expensive than selecting the same value of $x$ from dimension $i_{j+1}$ of the search space given that $x_{1\ldots k}$ are in the same normalised range and $\ j \in \mathbb{Z}^{+}_{k-1}$.
\label{def:CAC}
\end{def_preforder}
Based on Definition \ref{def:CAC}, cost-aware constraints are a sorted tuple of indexes that demonstrate the \textit{ordering} of dimensions of the search space based on the user's  knowledge about their cost of usage. 
Given a cost-aware preference as defined in Definition \ref{def:CAC} and a candidate solution such as $\mathbf x$, we formulate the cost of selecting $\mathbf x$ at iteration $t$ as a cost function:
\begin{equation}
{C}(\mathbf{x},t) = \prod_{ j=1}^{k} (1-\pi({x_{\mathcal{I}_{(j)}}},t))
\label{eq:cost}
\end{equation}
where $\mathcal{I}$ is a tuple with size of $k$, consist of cost-aware constraints orderings as defined in Definition \ref{def:CAC} and $\pi({x_{\mathcal{I}_{(j)}}},t)$ is sampled from an exponential distribution as:
\[
\pi({x_{\mathcal{I}_{(j)}}},t) \sim \mathrm{Exp}({x_{\mathcal{I}_{(j)}}},\lambda),\quad \lambda = \frac{1}{w_{{\mathcal{I}_{(j)}}}t+1}
\]
where $w_{\mathcal{I}_{(1)}},...,w_{\mathcal{I}_{(k)}} \sim {Dir}(1,...,1)$ such that $w_{\mathcal{I}_{(j)}} > w_{\mathcal{I}_{(j+1)}},\ \forall j \in \mathbb{Z}^{+}_{k-1}$. 
\begin{figure}
\begin{subfigure}[t]{0.49\linewidth}
    \centering
    \includegraphics[width=1\linewidth]{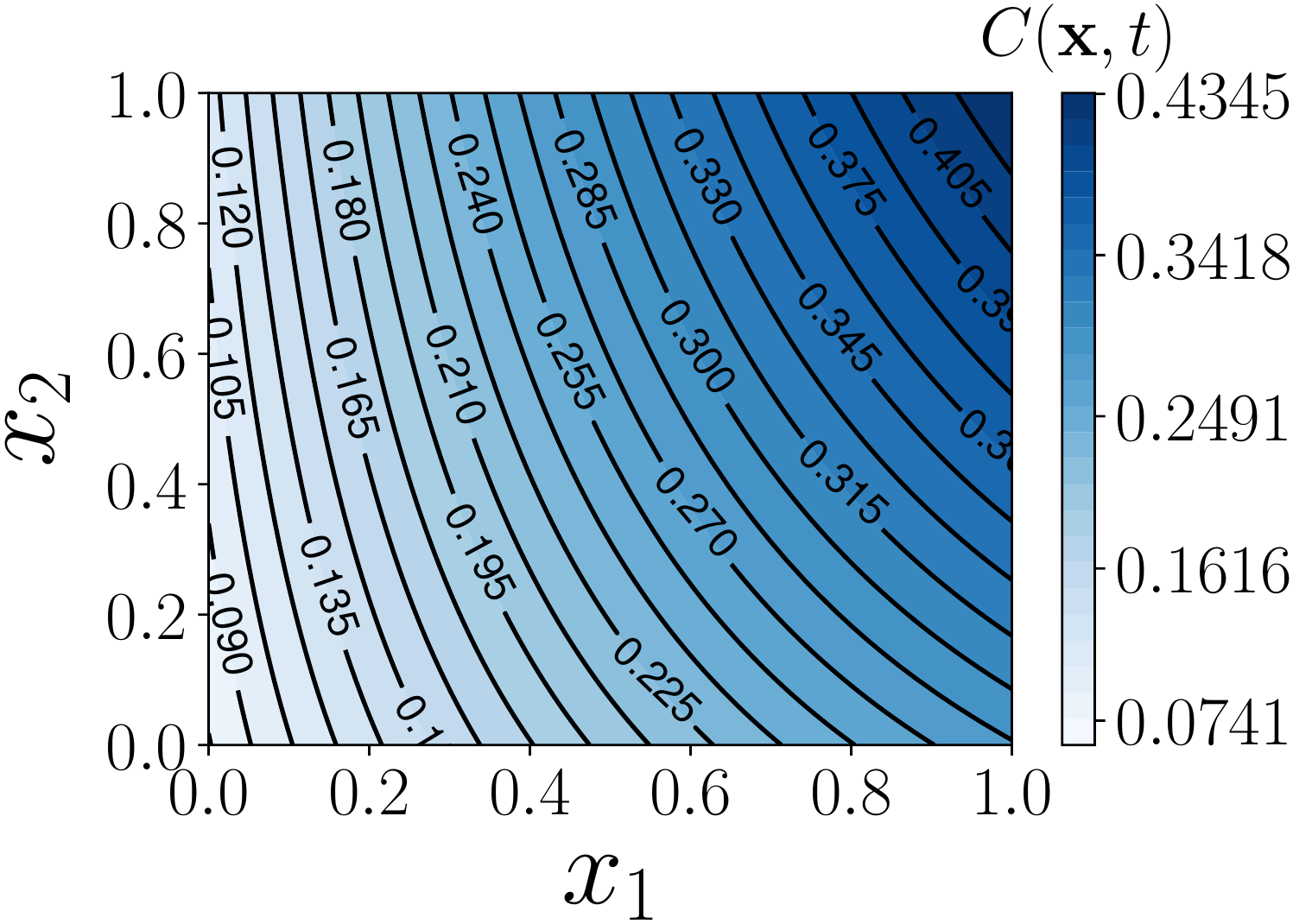}
    \caption{$t=1$}
    \label{fig:c1}
\end{subfigure}
\begin{subfigure}[t]{0.49\linewidth}
    \centering
    \includegraphics[width=1\linewidth]{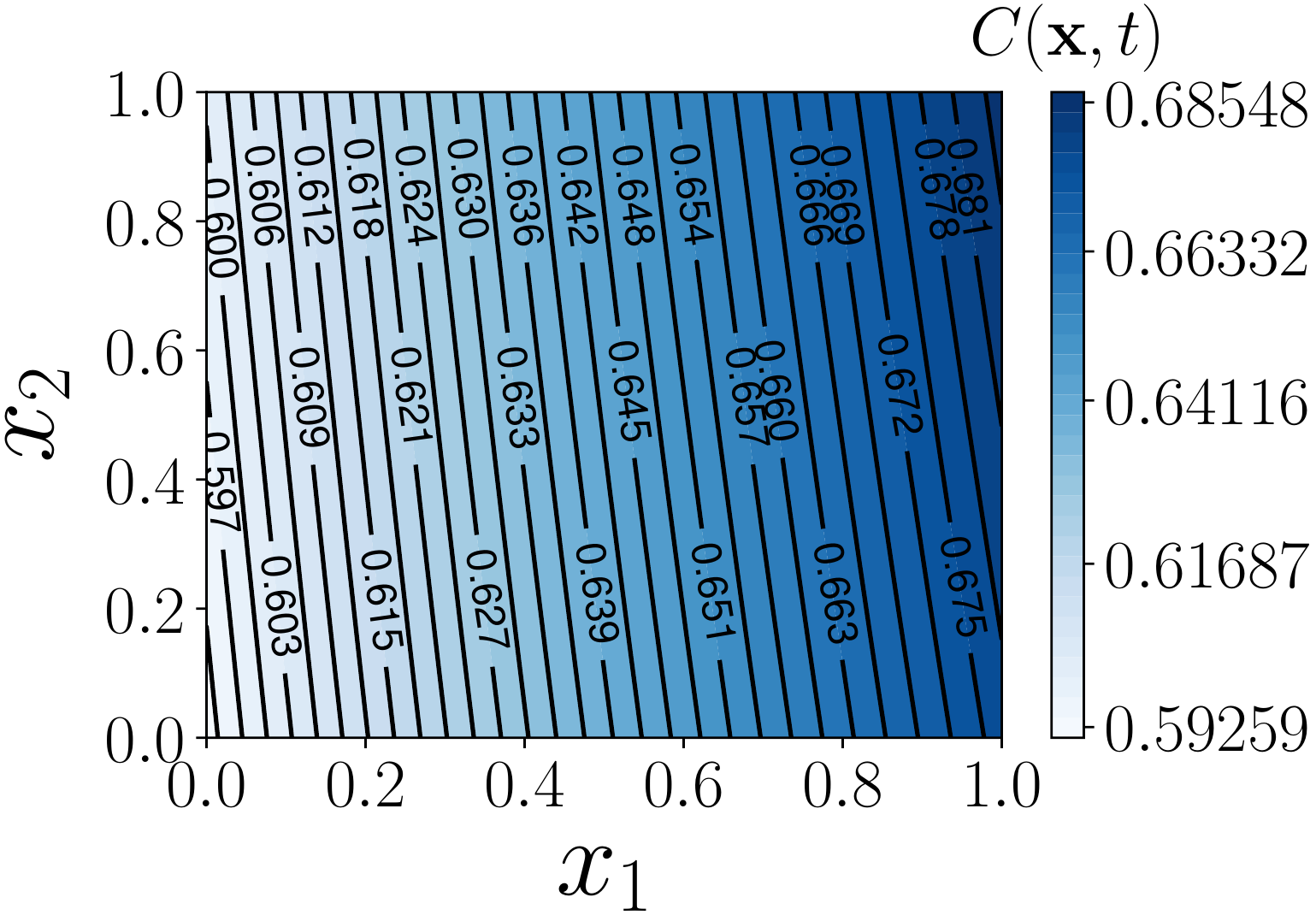}
    \caption{$t=10$}
    \label{fig:c2}       
\end{subfigure}
\begin{subfigure}[t]{0.49\linewidth}
    \centering
    \includegraphics[width=1\linewidth]{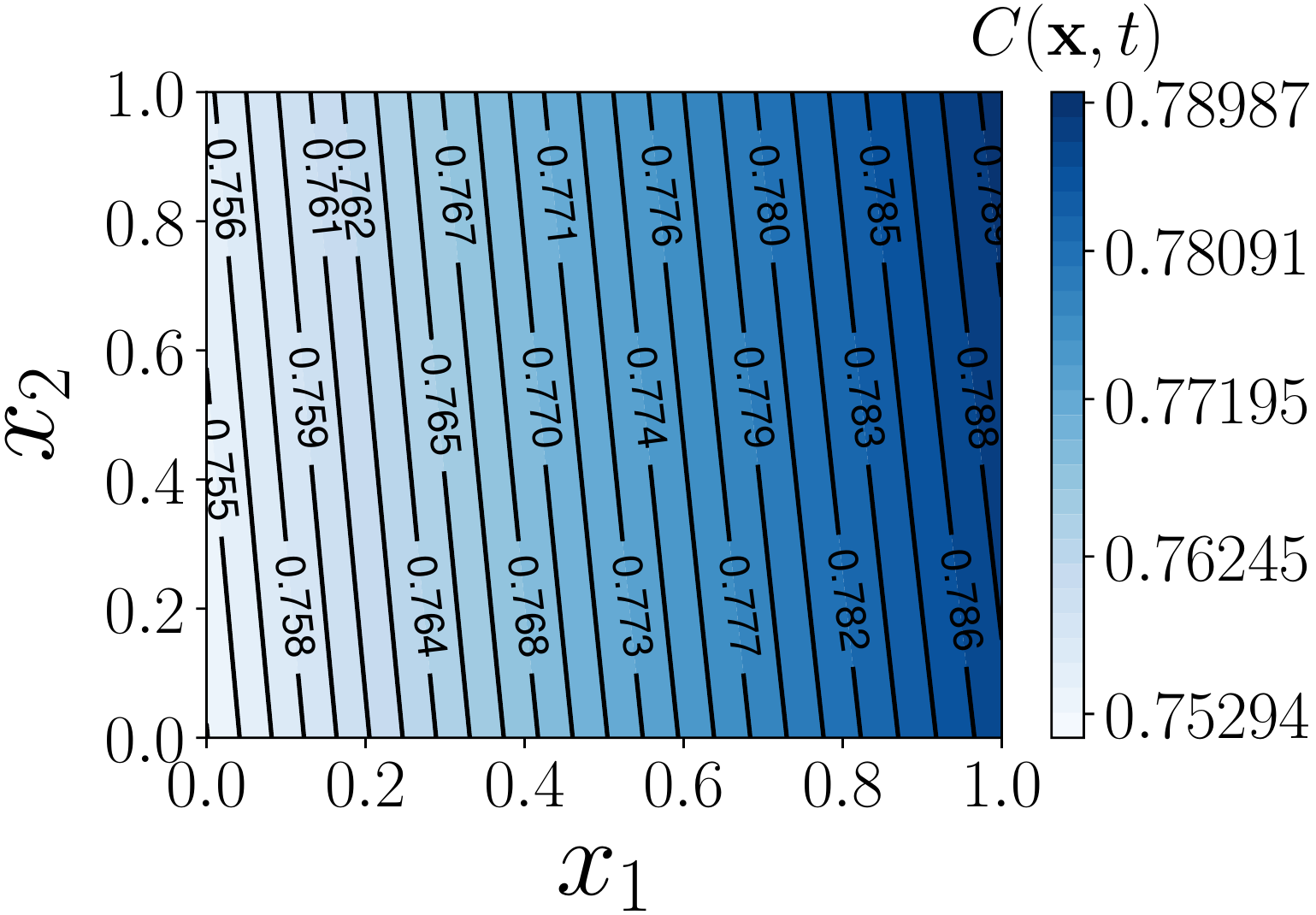}
    \caption{$t=20$}
    \label{fig:c3}      
\end{subfigure}
\begin{subfigure}[t]{0.49\linewidth}
    \centering
    \includegraphics[width=1\linewidth]{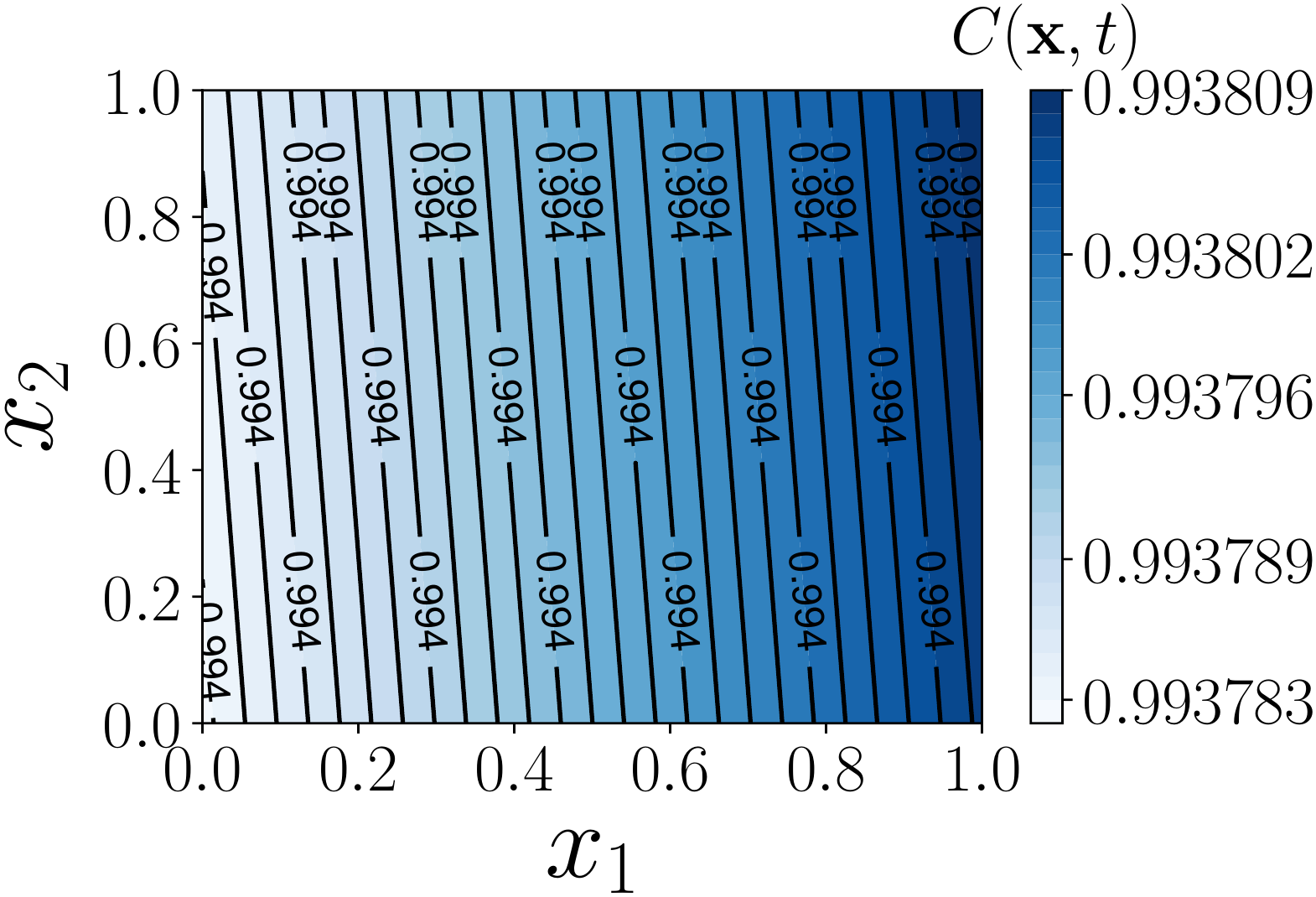}
    \caption{$t=1000$}
    \label{fig:c4}
\end{subfigure}
\caption{Illustration of the cost function with cost-aware constraint $\mathcal{I} = (1,2)$. Figure \ref{fig:c1} shows the cost of selecting a combination of  $x_1$ and $x_2$ when $t=1$. When high values of $x_1$ and $x_2$ is selected, $C(\mathbf{x},t)$ is significantly higher than the cost of low values of $x_1$ and $x_2$. As the optimisation progress, $C(\mathbf{x},t)$  increases for any combination of $x_1$ and $x_2$ (Figure \ref{fig:c2} and \ref{fig:c3}) and the difference of cost for a cheap and expensive combination reduces. As for Figure \ref{fig:c4}, $C(\mathbf x,t)$ is close to $1$ for all the combinations.}
\label{fig:Cost}
\end{figure}
\indent{} \par
The idea behind the $C(\mathbf x,t)$ comes from the natural properties of exponential distribution in modeling situations where certain events occur with a constant probability per unit length. Given that the cheaper regions of the search space can be selected more often than the expensive regions, the cost function constructs different exponential distributions based on each dimension of the search space with respect to their ordering in cost-aware preferences. This is achieved  by using different $\lambda$ values.
 Having a higher cost of usage for a dimension of search space will result in a higher value of $w_{\mathcal{I}_{(k)}}$ and accordingly smaller values of $\lambda$ for its corresponding distribution for that dimension which reshapes the distribution based on cost-aware constraints. 
 $C(\mathbf x,t)$ follows our motivations introduced before and it is also independent from objective space since cost-aware preference has been defined uniquely on the search space and $t$. 
\indent{} \par 
Consider the example of $f:\mathbb{R}^2 \rightarrow \mathbb{R}^2$ and $\mathcal{I}=(1, 2)$, i.e. using dimension $1$ of the search space is more expensive than second dimension. Since the cost function is independent of $\mathbf f$ (objective space), we plot the cost function ${C}(\mathbf{x},t)$ with respect to different values of ${x_1}$, $x_2$ and $t$.
Figure \ref{fig:Cost} confirms that at $t=1$, there is a significant difference between the cost of a candidate point with an expensive combination of inputs (i.e. $x_1=0.9$ and $x_2=0.9$)
and the cost of a cheaper combination of the inputs from the search space (i.e. $x_1=0.1,x_2=0.2$). As the optimisation progresses, this gap will shrink and the effects of cost-aware constraints will diminish (see Figure \ref{fig:c2} and Figure \ref{fig:c3}). Finally, when $t \rightarrow \infty,\ C(\mathbf{x},t) \approx 1,\ \ \forall \mathbf{x}$ (see Figure \ref{fig:c4}).
\indent{} \par
We now define CA-MOBO cost-aware acquisition function based on scalarised UCB and the cost-aware constraints incorporated in a separate cost function as:
\begin{equation}
\alpha(\mathbf{x},{\bm \theta}_t,t) = {Q}(\mathbf{x},{\bm \theta}_t)\times (1-{C}(\mathbf{x},t))
\label{eq:AcqFunc}
\end{equation}
where ${Q}(\mathbf{x},{\bm \theta}_t)$ is the scalarised UCB as defined in (\ref{eq:Q}) and ${C}(\mathbf{x},t)$ calculates the cost of selecting $\mathbf x$ as the input at iteration $t$ (see  (\ref{eq:cost})).
Algorithm \ref{Alg:1} details the cost-aware multi-objective Bayesian Optimisation.
\begin{algorithm}[t]
\caption{CA-UCB}
\label{alg:sperp_test}
\begin{algorithmic}
 \State \textbf{Input:\ \ } 
 Initial observations $\mathbb{D}$, GP prior {${\mu}^{t=1} ({\bf x}) = 0$}, $K$
 \For{$t=1,...,T$}
 \State Sample $\bm \theta_t \sim U_{\bm \theta_t}$ \Comment{{\small \textit{for Eq (\ref{eq:cheb})}}} 
 \State Find ${\bf x}_t = \argmaxx_{{\bf x} \in \mathbb{X}}\ \ \alpha({\bf x},{\bm \theta}_t,t)$  \Comment{{\small \textit{defined in Eq (\ref{eq:AcqFunc})}}}
 \State Update $\mathbb{D} \cup \{{\bf x},f({\bf x})\}$ 
 \State Update $\mathcal{GP}_m\ \ \ \forall m \in \mathbb{Z}^{+}_M$
 \EndFor
 \end{algorithmic}
 \label{Alg:1}
\end{algorithm} 

\section{Theoretical Bounds}
We first define our proposed instantaneous and cumulative regret for a cost-aware multi-objective optimisation problem and then derive 
the theoretical upper bound on the cumulative regret. 
The proof builds on the ideas in \cite{srinivas2009gaussian} and  recently published work in multi-objective optimisation \cite{Kand2018}. In this section we assume that the kernel hyperparameters are known. 
\indent{} \par
The regret defined in this problem must be representative of both compliance with cost-aware constraints and also the goodness of the Pareto front.
The instantaneous regret (simple regret) incurred by CA-MOBO at iteration number $t$ is:
\begin{align}
r(\mathbf{x}_t,{\bm \theta}_t) = \mathrm{\maxx_{{\bf x} \in \mathcal{X}}}\ \  S_{{\bm \theta}_t} \big(f({\bf x})\big)\big(1-C({\bf x},t)\big) - ...\\\nonumber
...S_{{\bm \theta}_t} \big(f({\bf x}_t)\big)\big(1-C({\bf x}_t,t)\big)
\label{eq:Reg}
\end{align}
where:
$
{\bf x}_t = \argmax_{{\bf x} \in \mathbb{X}}\ \  \alpha({\bf x},{\bm \theta}_t,t).
$
${\bm \theta}_t$ is the sampled weights for scalarisation function $S$ based on an arbitrary distribution $U_{\bm \theta}$ as defined in (\ref{eq:cheb})  and $C({\bf x},t)$ is the cost function for input ${\bf x}$ at iteration $t$ as defined in (\ref{eq:cost}). Accordingly the cumulative regret is calculated as:
\begin{align}
\mathcal{R}(T) = \sum_{t=1}^{T} r(\mathbf{x}_t,{\bm \theta}_t) = \sum_{t=1}^{T} \Big(\mathrm{\maxx_{{\bf x} \in \mathcal{X}}}\ \  S_{{\bm \theta}_t} \big(f({\bf x})\big)...\\ \nonumber...\big(1- C({\bf x},t)\big)-S_{{\bm \theta}_t} \big(f({\bf x}_t)\big)\big(1-C({\bf x}_t,t)\big)\Big)\label{eq:CR}
\end{align}
\begin{th_geomofS} 
Given the cost-aware acquisition function defined in (\ref{eq:AcqFunc}) with $\beta_t = 2ln(\frac{t^2|\mathbb{X}|}{\sqrt{2\pi}})$, the cumulative regret $\mathcal{R}(T)$ is upper bounded as:
\[
\bar{U}_{\bm \theta}\Big(MT\beta_T \sum_{m=1}^{M} \frac{\gamma_{T(m)}}{ln(1+\sigma^{-2}_{m})}\Big)^{\frac{1}{2}} + \frac{\pi^2}{3}M \mathbb{E}[{U}_{\bm \theta}]
\]
where ${\gamma_{T(m)}}$ is the maximum information gain after $T$ iteration for objective function $m$ as defined in \cite{srinivas2009gaussian}. ${\bar{U}}_{\bm \theta}$ is also defined as ${\bar{U}}_{\bm \theta} = \mathbb{E} \big[ \sqrt{\frac{1}{T} \sum_{t=1}^{T} U^2_{{\bm \theta}_t}} \big]$.
\label{Th:CReg}
\end{th_geomofS} 
\begin{proof}
The proof is provided in the supplementary materials.
\end{proof}
\indent{} \par
Based on the Theorem \ref{Th:CReg}, we can see that the upper bound on the regret for CA-MOBO is no worse (in order) than the cumulative regret bound introduced in \cite{Kand2018} even though cost-aware constraints restricts exploration during early iterations.

\section{Experiments}
We now present our experimental results comparing the performance of CA-MOBO to other strategies without cost awareness.  These experiments including multiple synthetic and two real-world problems of optimising  the  hyperparameters of a feed-forward neural network and a random forest. For all synthetic functions, the experiment was repeated $50$ times with $500$ iterations and we map the search space and objective space to $[0,1]$. For real-world experiments, we report the average values of $10$ runs in $300$ iterations and the initial observations are randomly selected. The hyperparameters of the GP are updated based on the observed data every $10$ evaluations. Squared Exponential (SE) kernel is used in all experiments.
\indent{} \par
To the best of our knowledge there are no studies aiming to solve our proposed problem.  However, we compare our results with \cite{Kand2018}, as a scalarised multi-objective UCB (MO-UCB) with no cost-awareness. 
Additional experiments and CA-MOBO source codes are available in supplementary materials.
\begin{figure*}
\begin{subfigure}[t]{0.255\linewidth}
    \centering
    \includegraphics[scale=0.324]{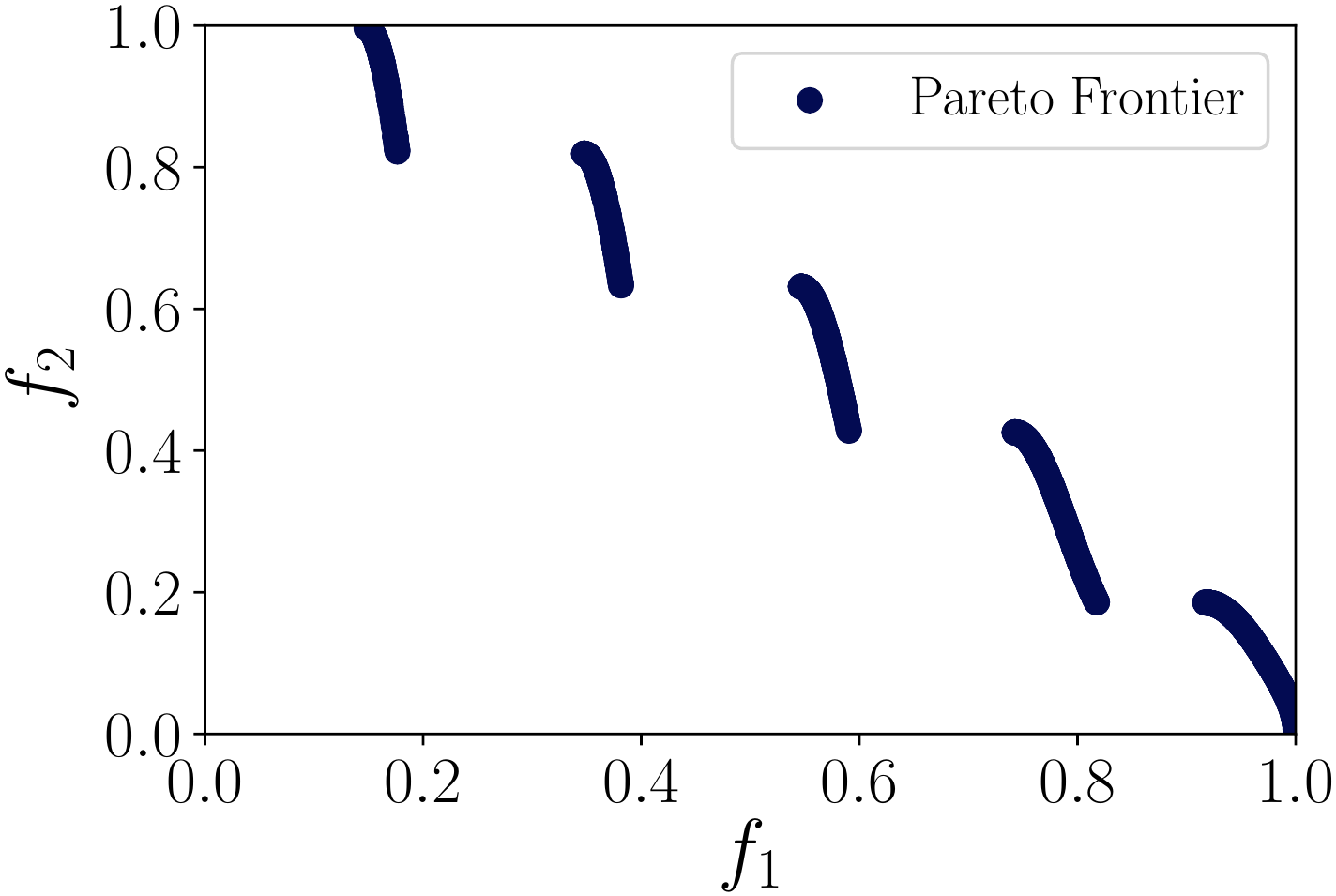}
    \caption{Full Pareto front}
    \label{fig:1x1}
\end{subfigure}
\begin{subfigure}[t]{0.245\linewidth}
    \centering
    \includegraphics[width=1\linewidth]{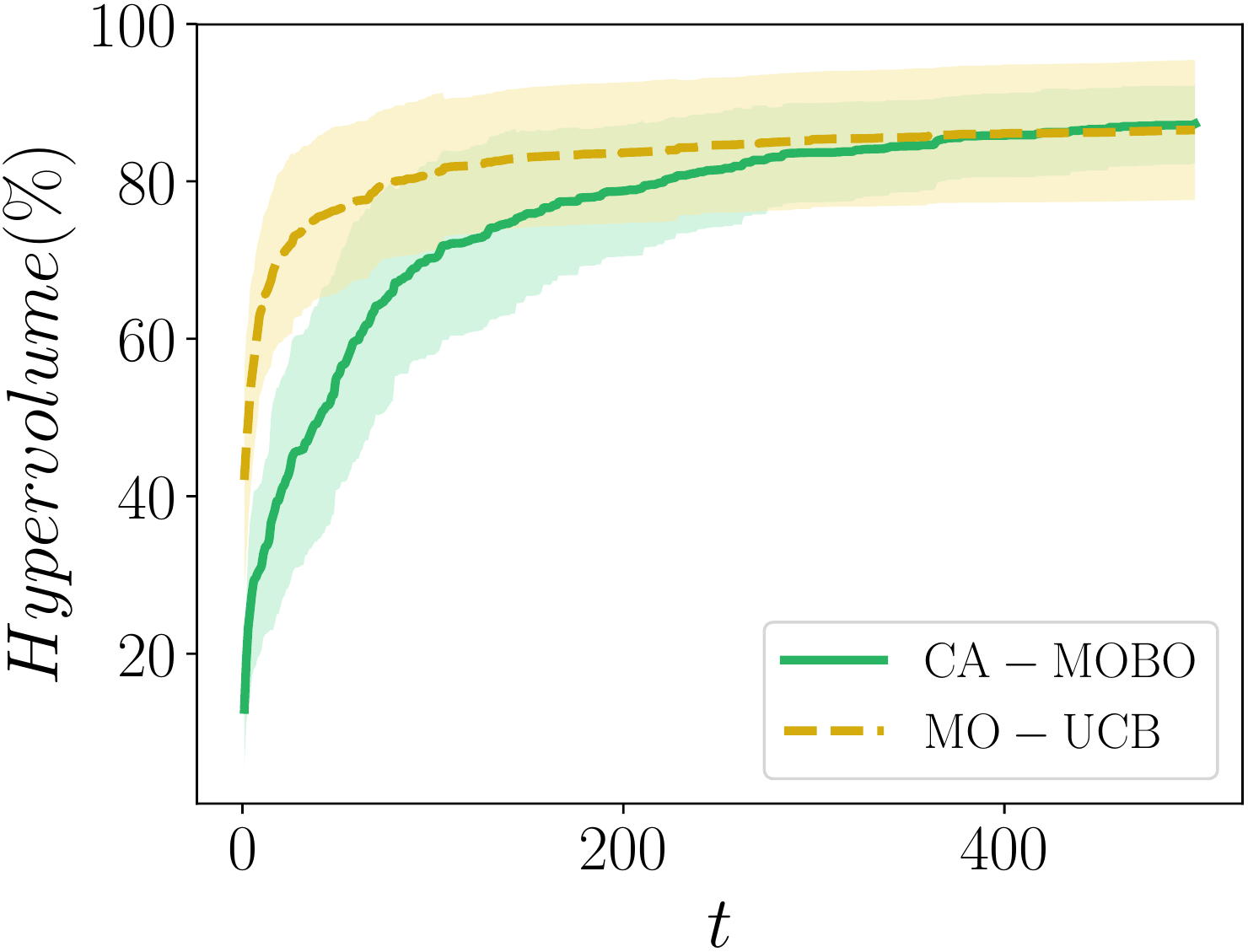}
    \caption{Dominated Hypervolume}
    \label{fig:1x2}
\end{subfigure}
\begin{subfigure}[t]{0.245\linewidth}
    \centering
    \includegraphics[width=1\linewidth]{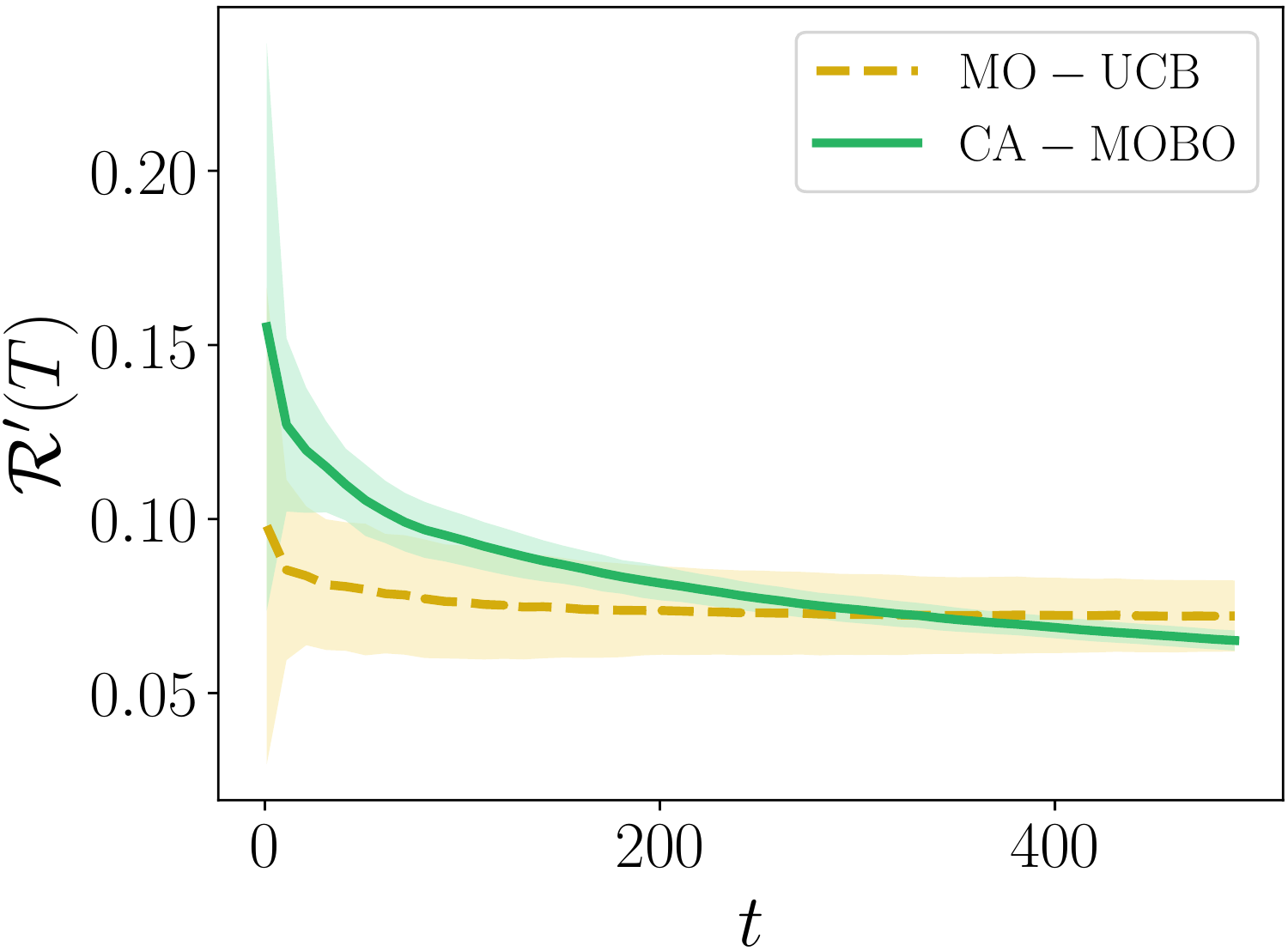}
    \caption{Average Regret}
    \label{fig:1x3}      
\end{subfigure}
\begin{subfigure}[t]{0.241\linewidth}
    \centering
    \includegraphics[width=1\linewidth]{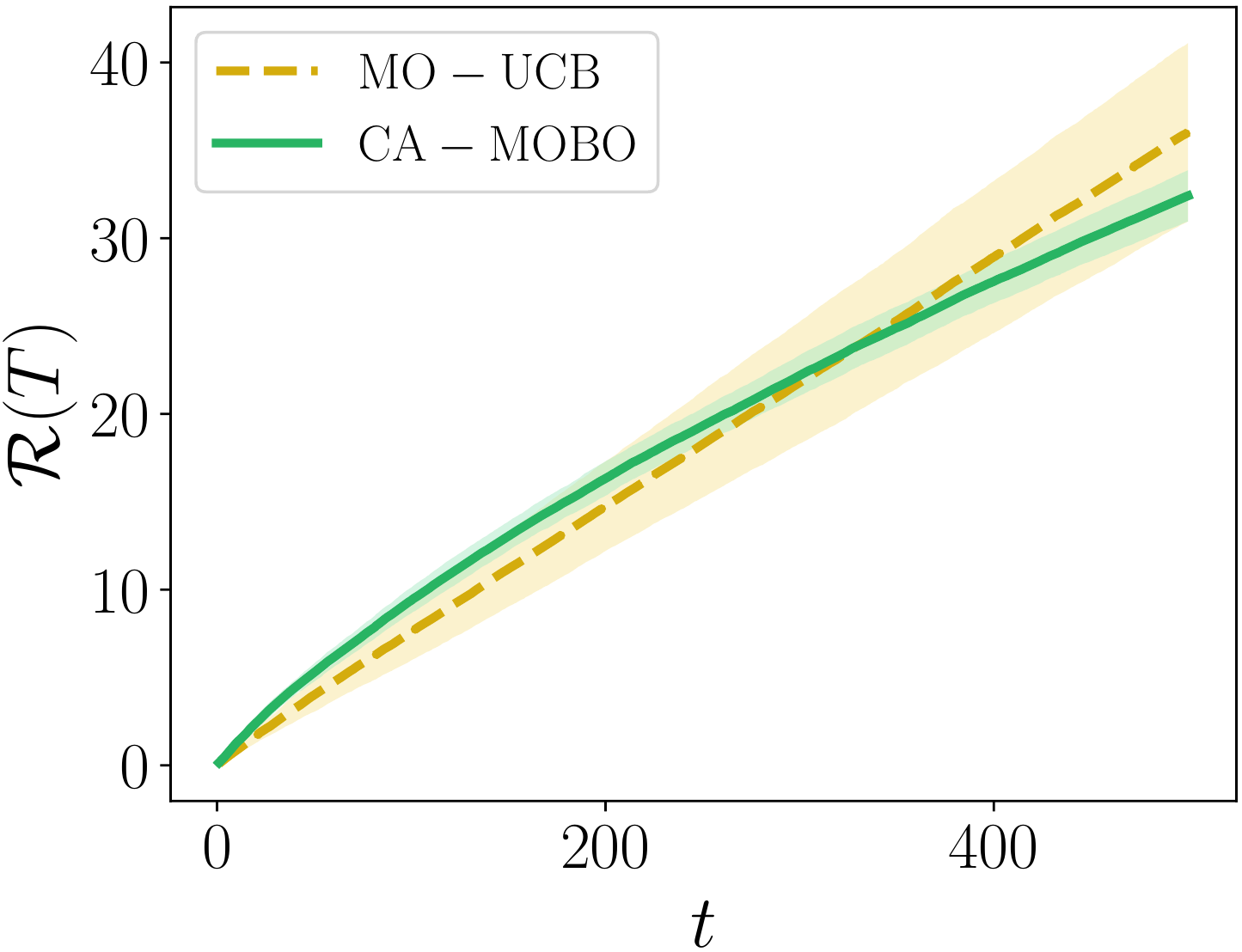}
    \caption{Cumulative Regret}
    \label{fig:1x4}
\end{subfigure}
\begin{subfigure}[t]{0.245\linewidth}
    \centering
    \includegraphics[width=1\linewidth]{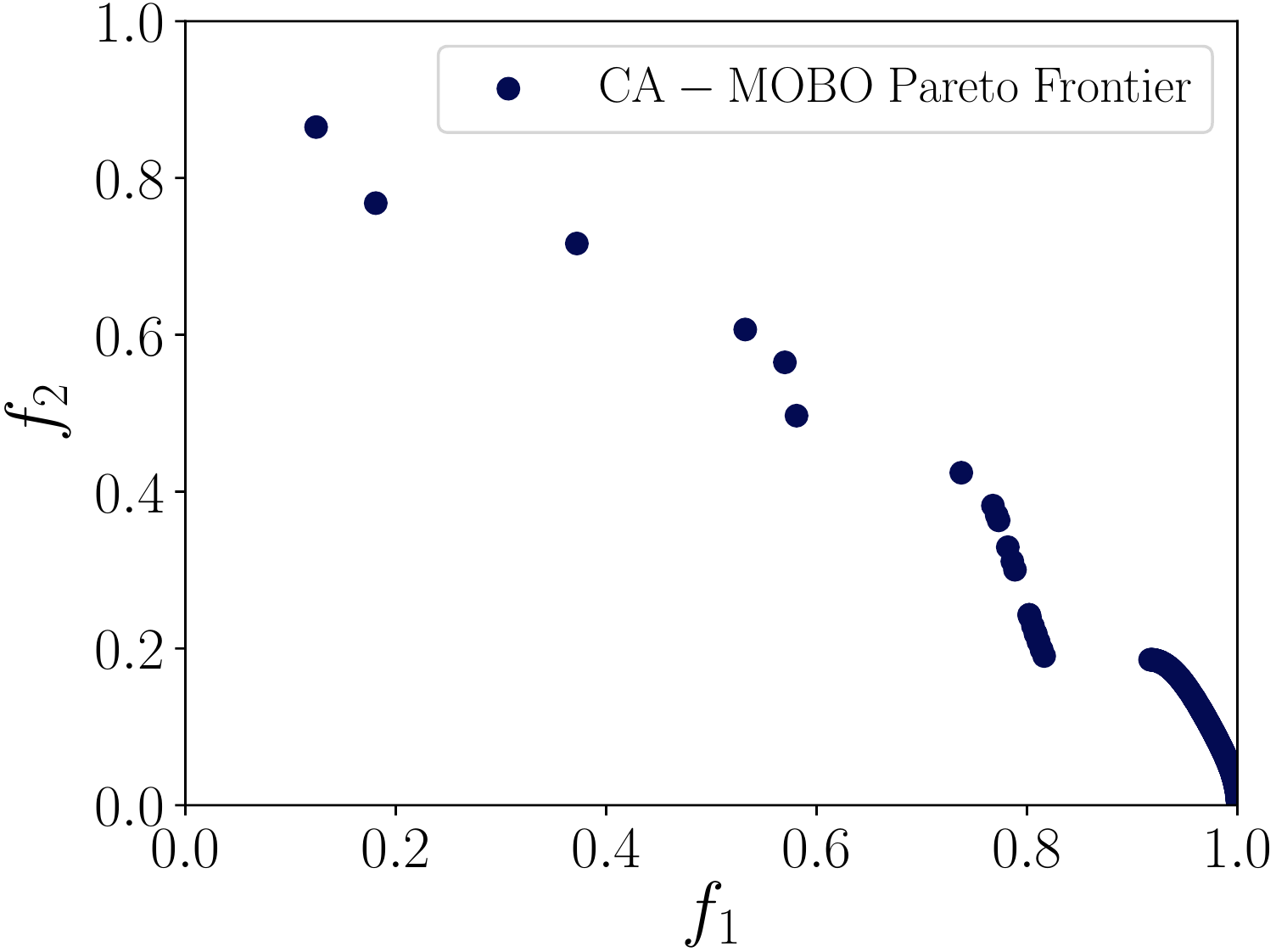}
    \caption{Pareto front (CA-MOBO) }
    \label{fig:1x5}       
\end{subfigure}
\begin{subfigure}[t]{0.245\linewidth}
    \centering
    \includegraphics[width=1\linewidth]{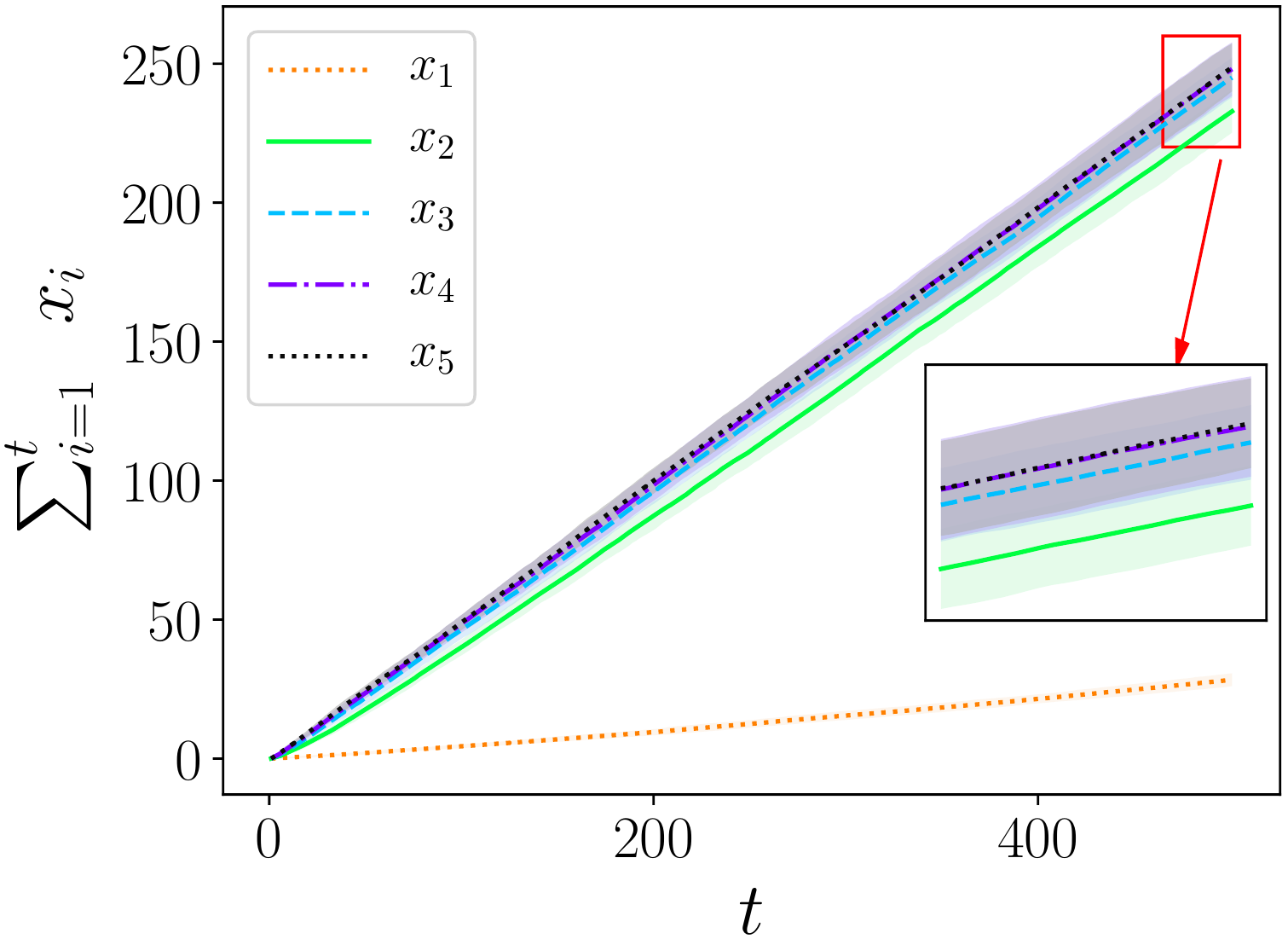}
    \caption{Sum of selected inputs (CA-MOBO)}
    \label{fig:1x6}      
\end{subfigure}
\begin{subfigure}[t]{0.245\linewidth}
    \centering
    \includegraphics[width=1\linewidth]{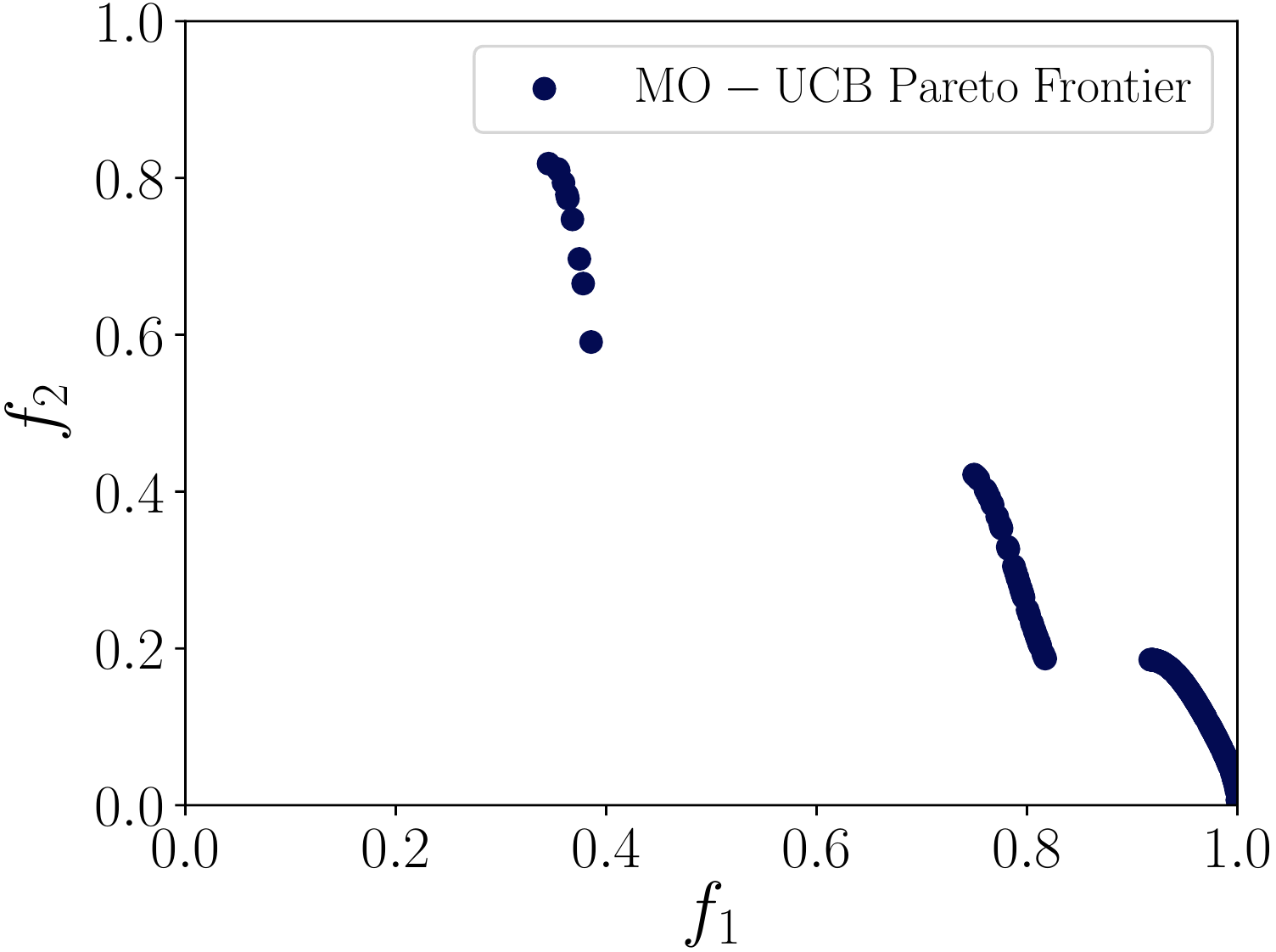}
    \caption{Pareto front (MO-UCB)}
    \label{fig:1x7}
\end{subfigure}
\begin{subfigure}[t]{0.25\linewidth}
    \centering
    \includegraphics[width=1\linewidth]{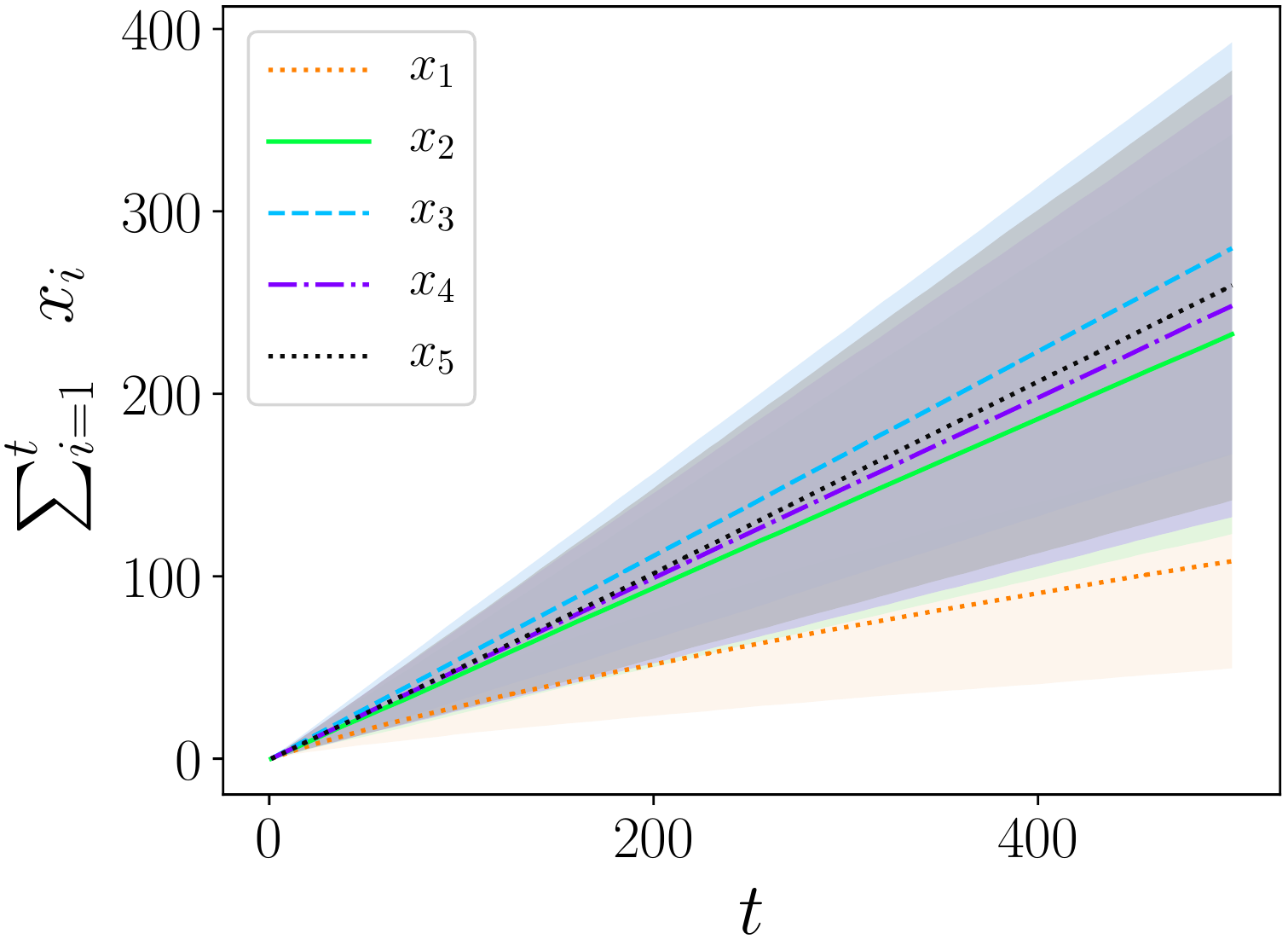}
    \caption{Sum of selected inputs (MO-UCB)}
    \label{fig:1x8}      
\end{subfigure}
\caption{Minimising Zitzler-Deb-Thiele's N. 3 \cite{zitzler2000comparison} function. Figure \ref{fig:1x1} shows the full Pareto front. Figure \ref{fig:1x2} shows the comparison of dominated hypervolume for CA-MOBO and MO-UCB.
Figure \ref{fig:1x3} illustrates the average regret of both methods. Figure \ref{fig:1x4} demonstrates the cumulative regret. Figure \ref{fig:1x5} illustrates the Pareto front obtained by CA-MOBO. Figure \ref{fig:1x6} shows the cumulative amount of selected inputs for each dimension of search space. 
Figure \ref{fig:1x7} and Figure \ref{fig:1x8} illustrate the obtained Pareto front  by MO-UCB and the corresponding cumulative amount of selected inputs respectively.}
\label{fig:exp1}
\end{figure*}

\begin{figure*}
\begin{subfigure}[t]{0.255\linewidth}
    \centering
    \includegraphics[scale=0.324]{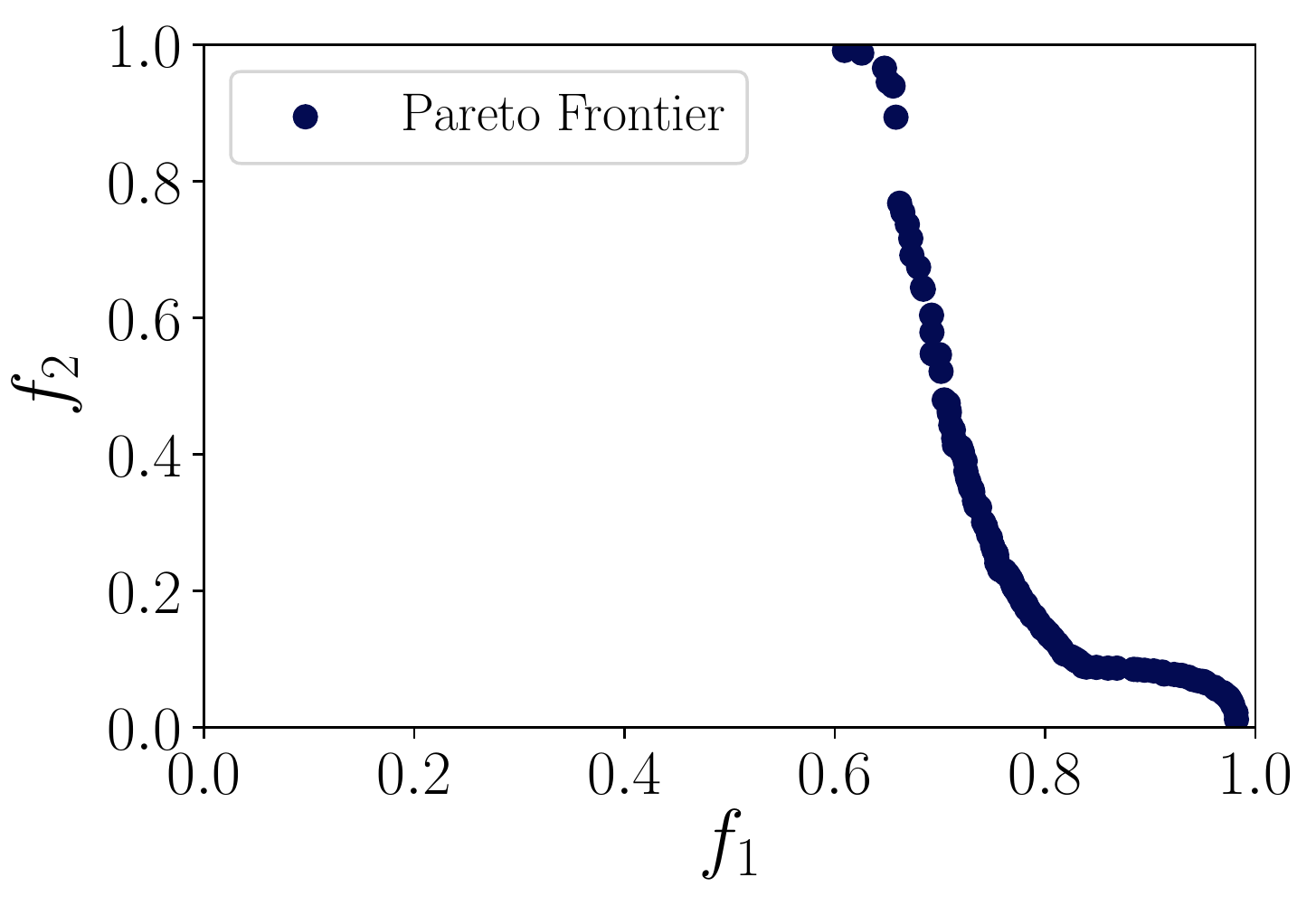}
    \caption{Full Pareto front}
    \label{fig:3x1}
\end{subfigure}
\begin{subfigure}[t]{0.245\linewidth}
    \centering
    \includegraphics[width=1\linewidth]{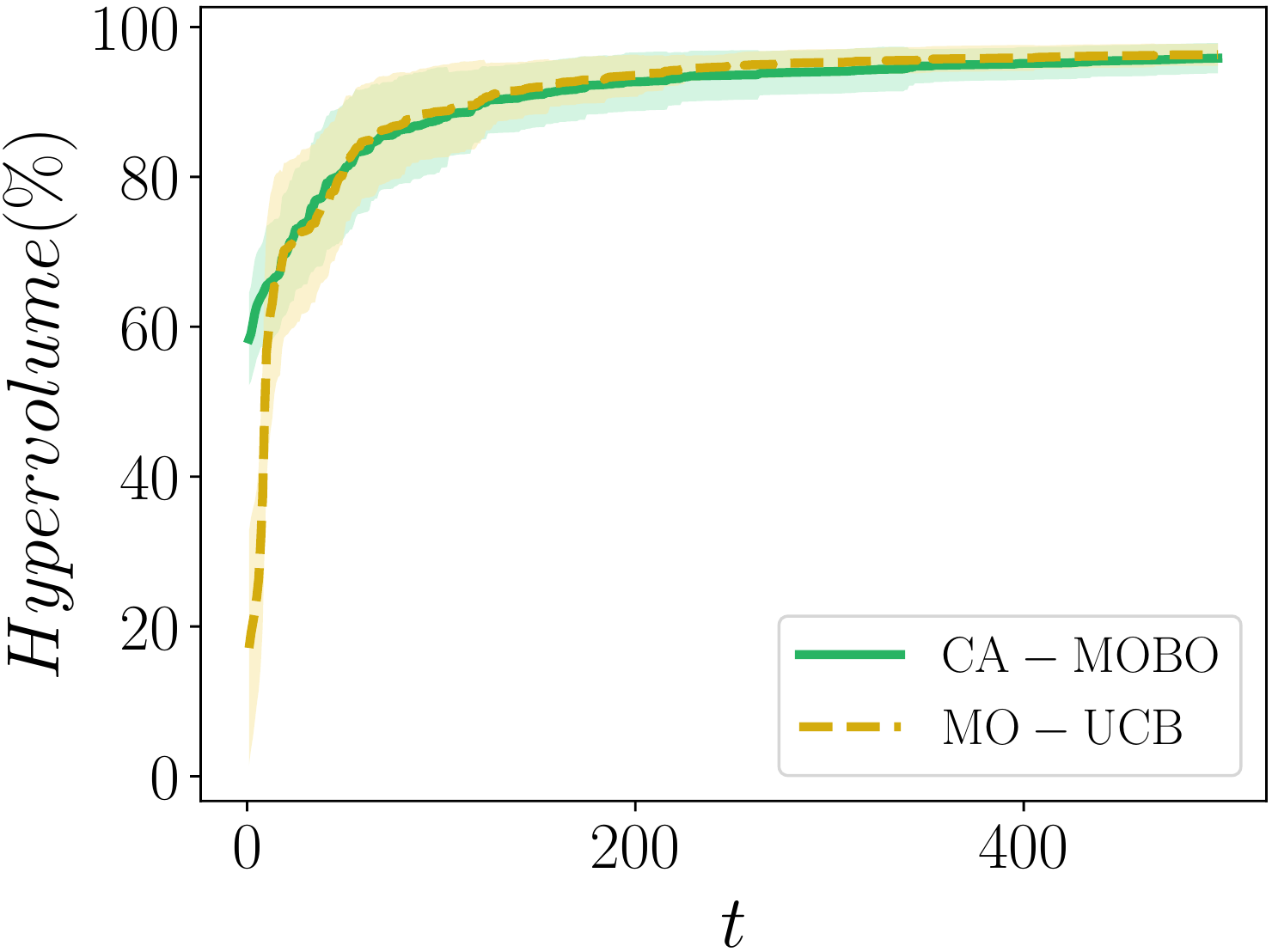}
    \caption{Dominated Hypervolume}
    \label{fig:3x2}
\end{subfigure}
\begin{subfigure}[t]{0.245\linewidth}
    \centering
    \includegraphics[width=1\linewidth]{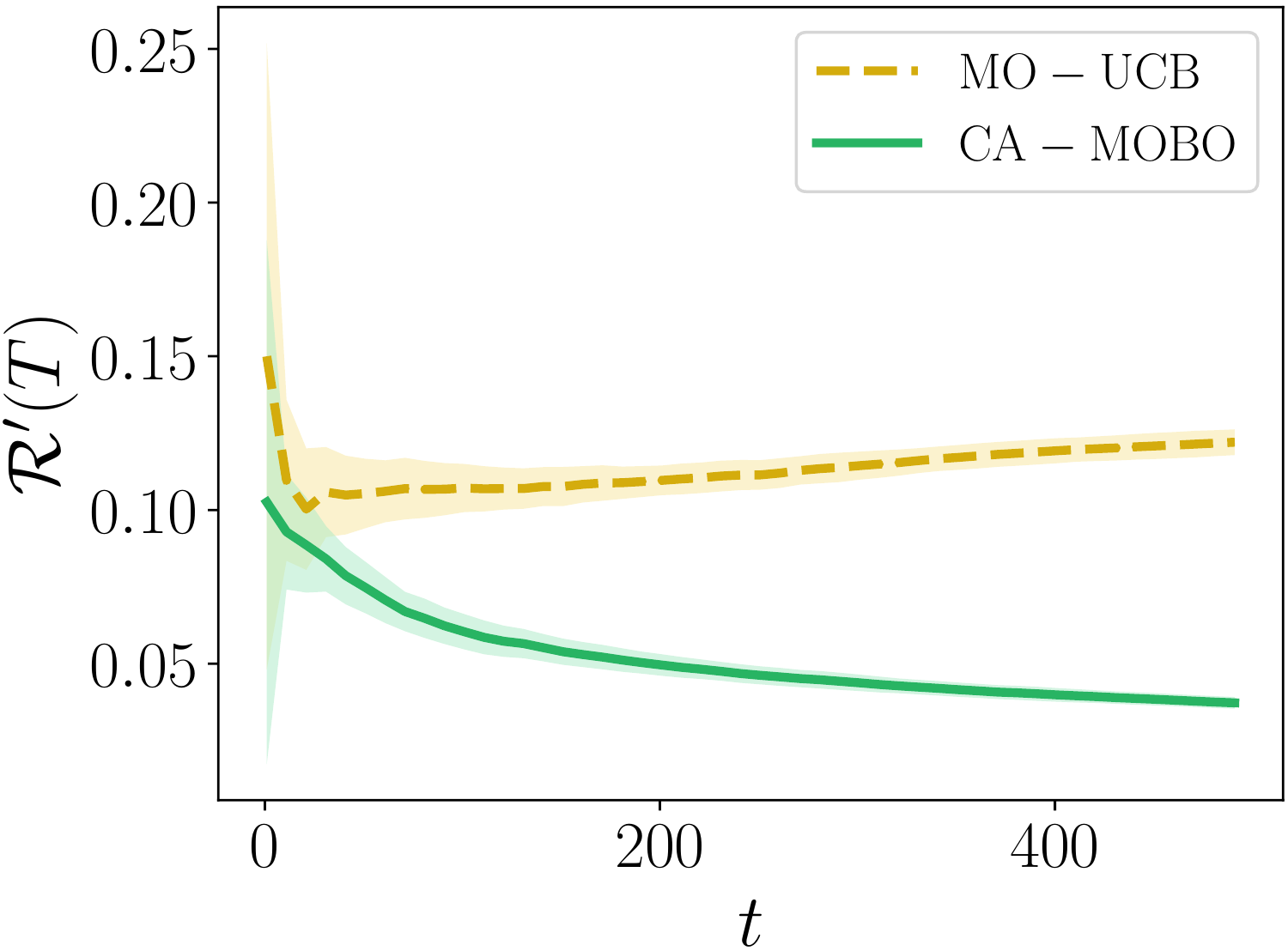}
    \caption{Average Regret}
    \label{fig:3x3}      
\end{subfigure}
\begin{subfigure}[t]{0.241\linewidth}
    \centering
    \includegraphics[width=1\linewidth]{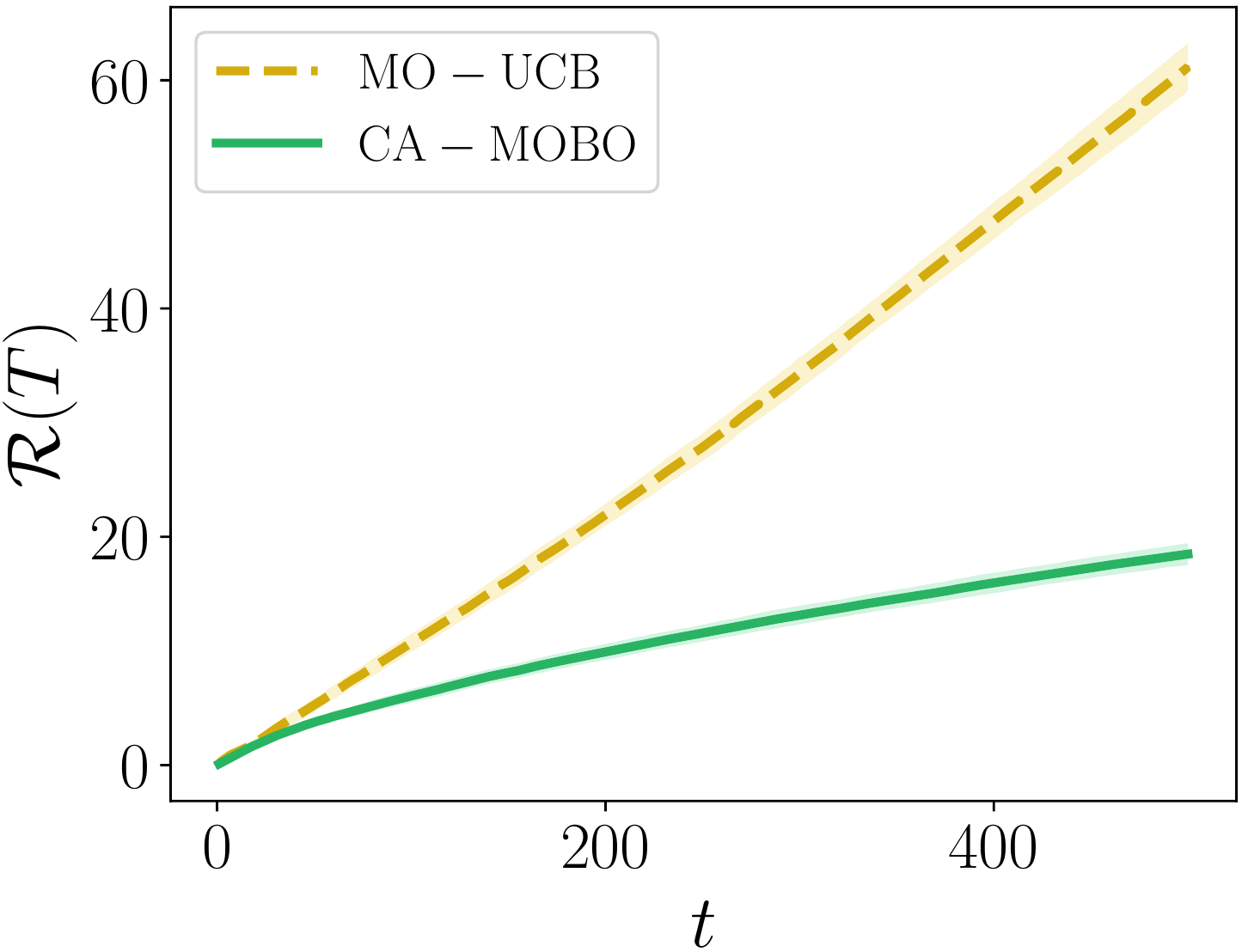}
    \caption{Cumulative Regret}
    \label{fig:3x4}
\end{subfigure}
\begin{subfigure}[t]{0.245\linewidth}
    \centering
    \includegraphics[width=1\linewidth]{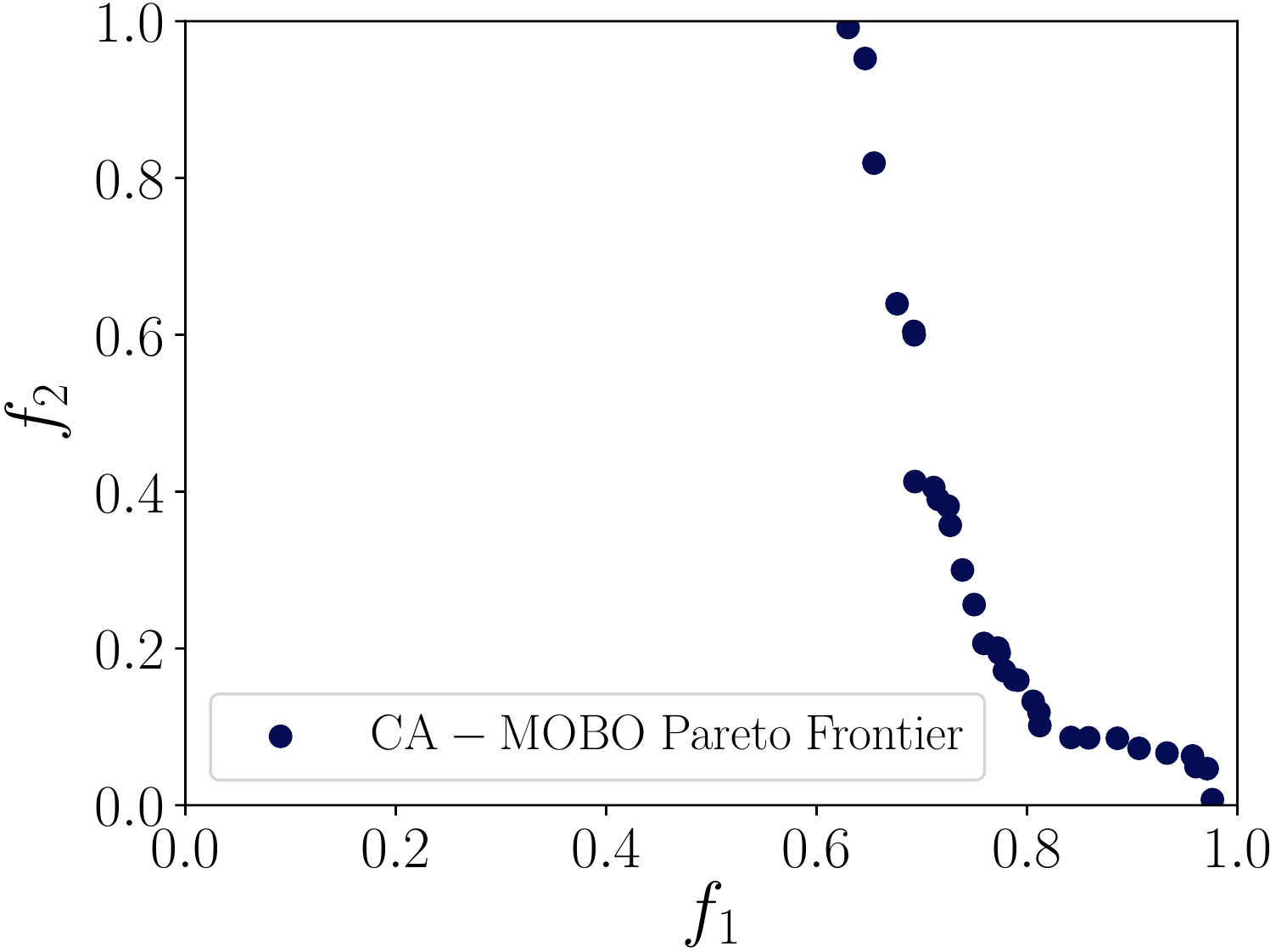}
    \caption{Pareto front (CA-MOBO) }
    \label{fig:3x5}       
\end{subfigure}
\begin{subfigure}[t]{0.245\linewidth}
    \centering
    \includegraphics[width=1\linewidth]{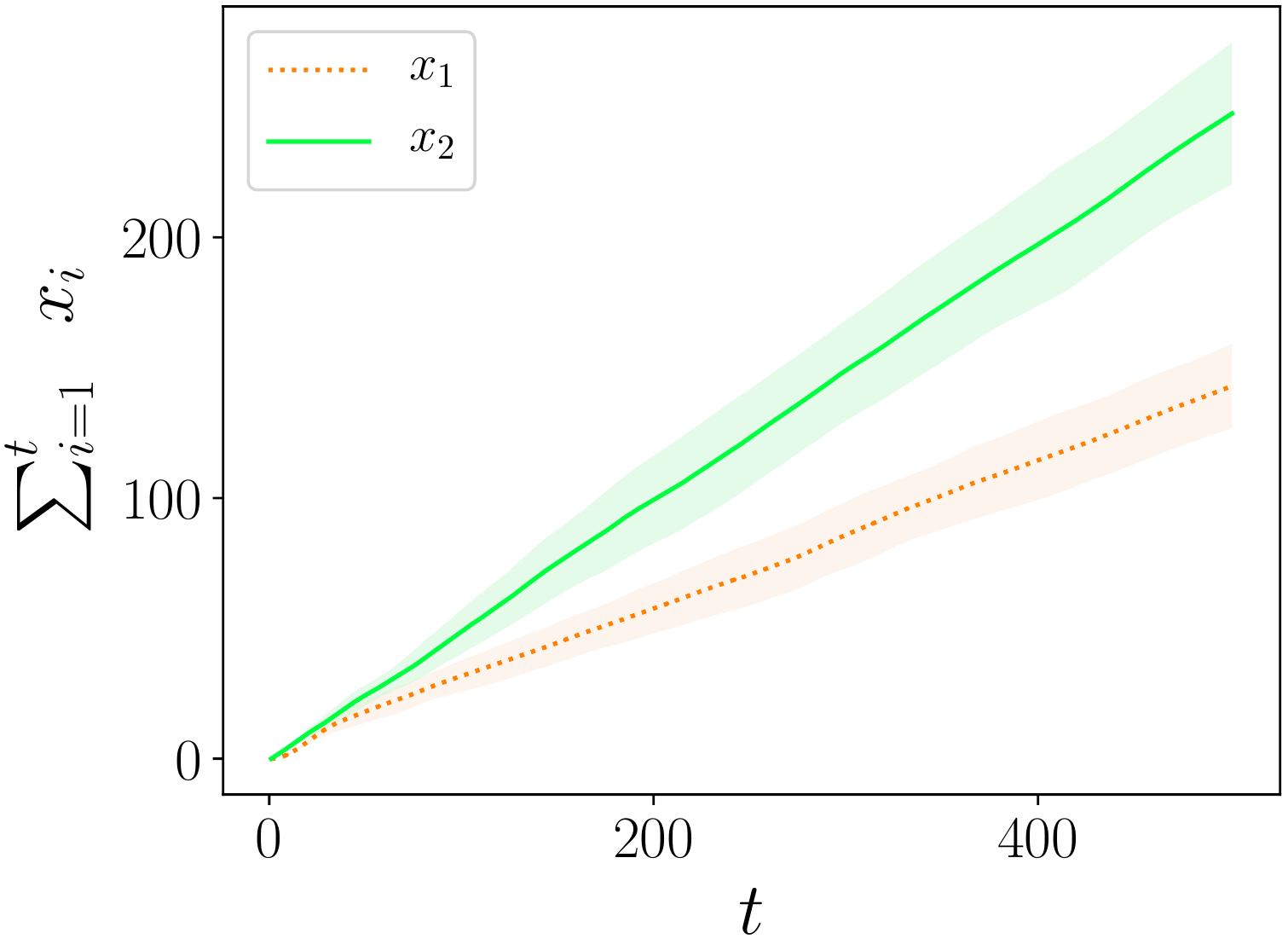}
    \caption{Sum of selected inputs (CA-MOBO)}
    \label{fig:3x6}      
\end{subfigure}
\begin{subfigure}[t]{0.245\linewidth}
    \centering
    \includegraphics[width=1\linewidth]{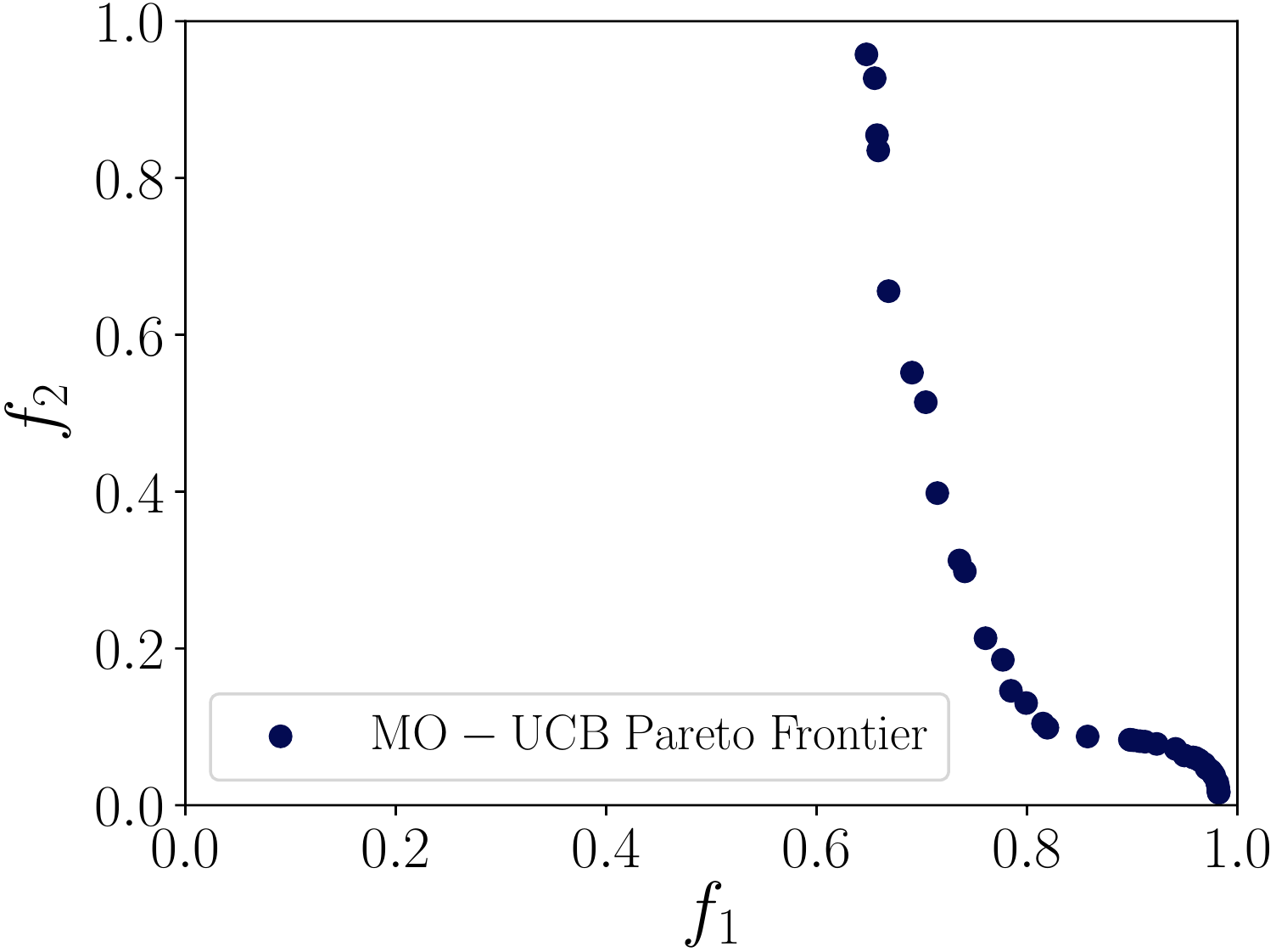}
    \caption{Pareto front (MO-UCB)}
    \label{fig:3x7}
\end{subfigure}
\begin{subfigure}[t]{0.25\linewidth}
    \centering
    \includegraphics[width=1\linewidth]{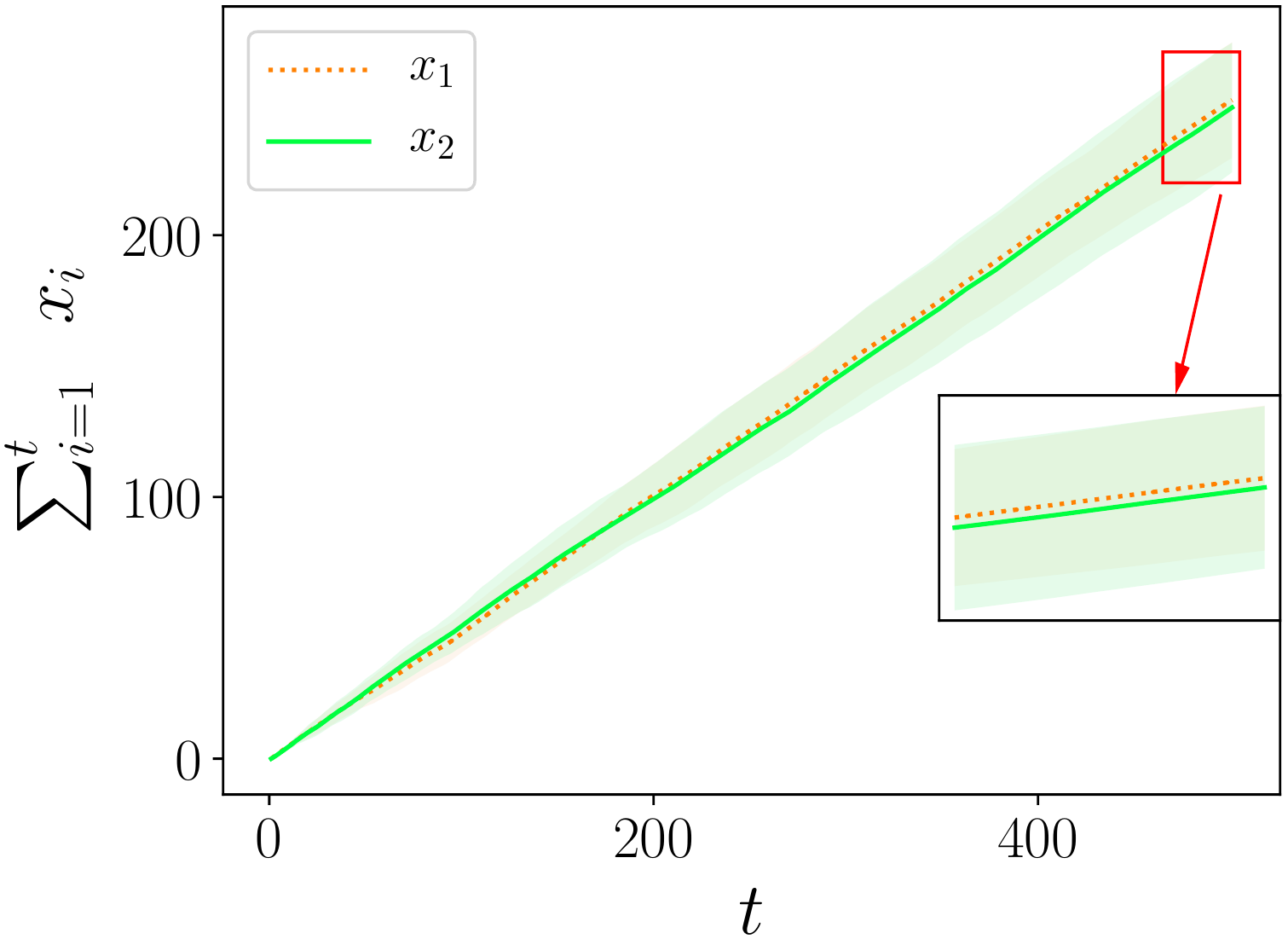}
    \caption{Sum of selected inputs (MO-UCB)}
    \label{fig:3x8}      
\end{subfigure}
\caption{Minimising Cross-in-tray function as first objective and Hölder table function and the second objective. Figure \ref{fig:3x1} shows the full Pareto front. Figure \ref{fig:3x2} demonstrates a comparison of the dominated hypervolume for MO-UCB and CA-MOBO. Figure \ref{fig:3x3} and Figure \ref{fig:3x4} demonstrate the comparison of the average regret and cumulative regret for both methods respectively. Figure \ref{fig:3x5} and Figure \ref{fig:3x7} show Obtained Pareto front by CA-MOBO and MO-UCB, respectively. Figure \ref{fig:3x6} shows the calculation of $\sum_{i=1}^{T}\ x_i$ as the sum of selected inputs  for CA-MOBO and it confirms the compliance of CA-MOBO with the cost-aware constraints. Figure \ref{fig:3x8} illustrates  sum of selected inputs for MO-UCB.}
\label{fig:exp3}
\end{figure*}
\begin{figure*}
\begin{subfigure}[t]{0.24\linewidth}
    \centering
    \includegraphics[width=1\linewidth]{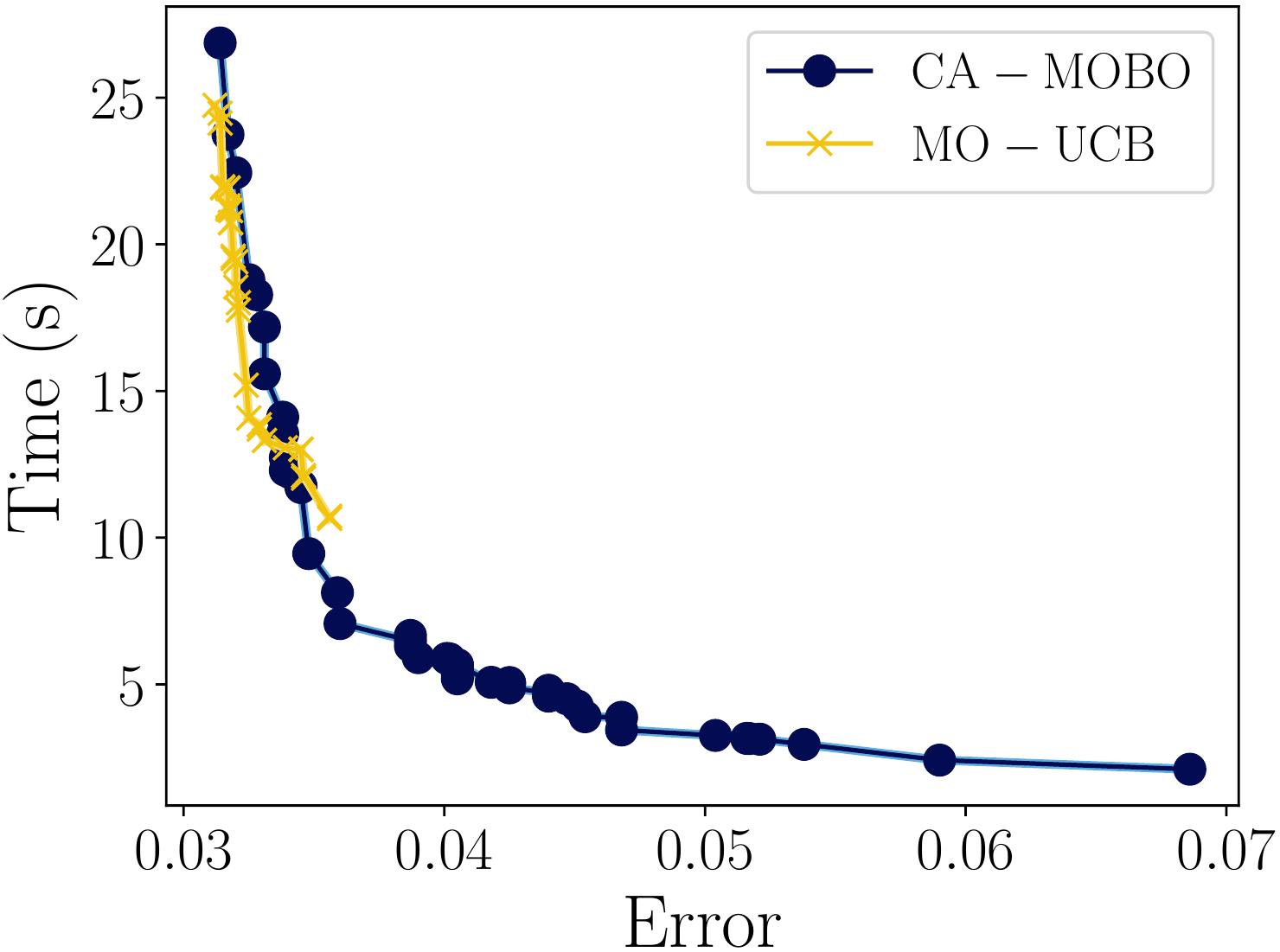}
    \caption{Obtained Pareto front}
    \label{fig:5x1}
\end{subfigure}
\begin{subfigure}[t]{0.24\linewidth}
    \centering
    \includegraphics[width=1\linewidth]{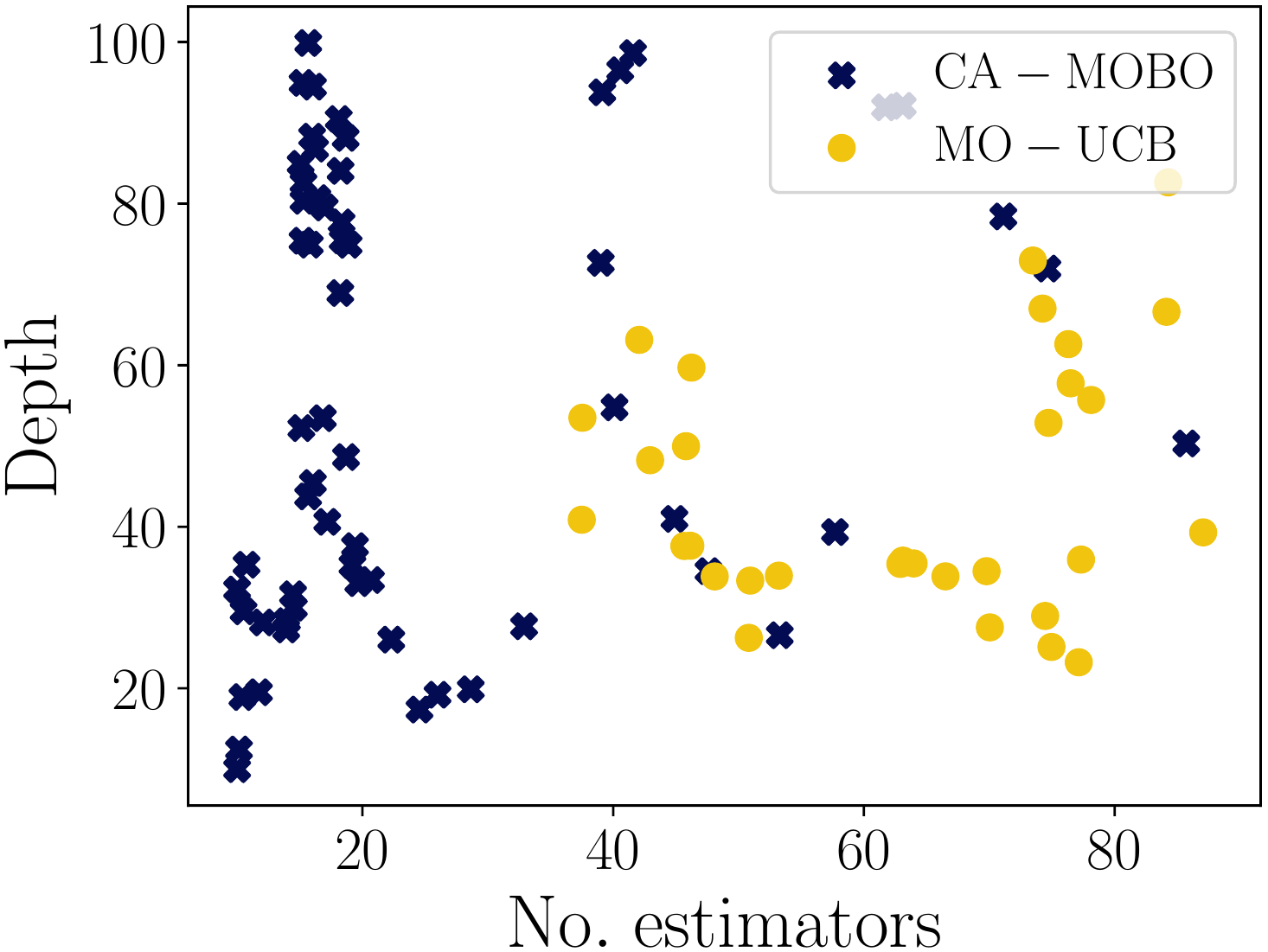}
    \caption{Pareto fronts solutions found in $\mathbb{X}$}
    \label{fig:5x2}       
\end{subfigure}
\begin{subfigure}[t]{0.25\linewidth}
    \centering
    \includegraphics[width=1\linewidth]{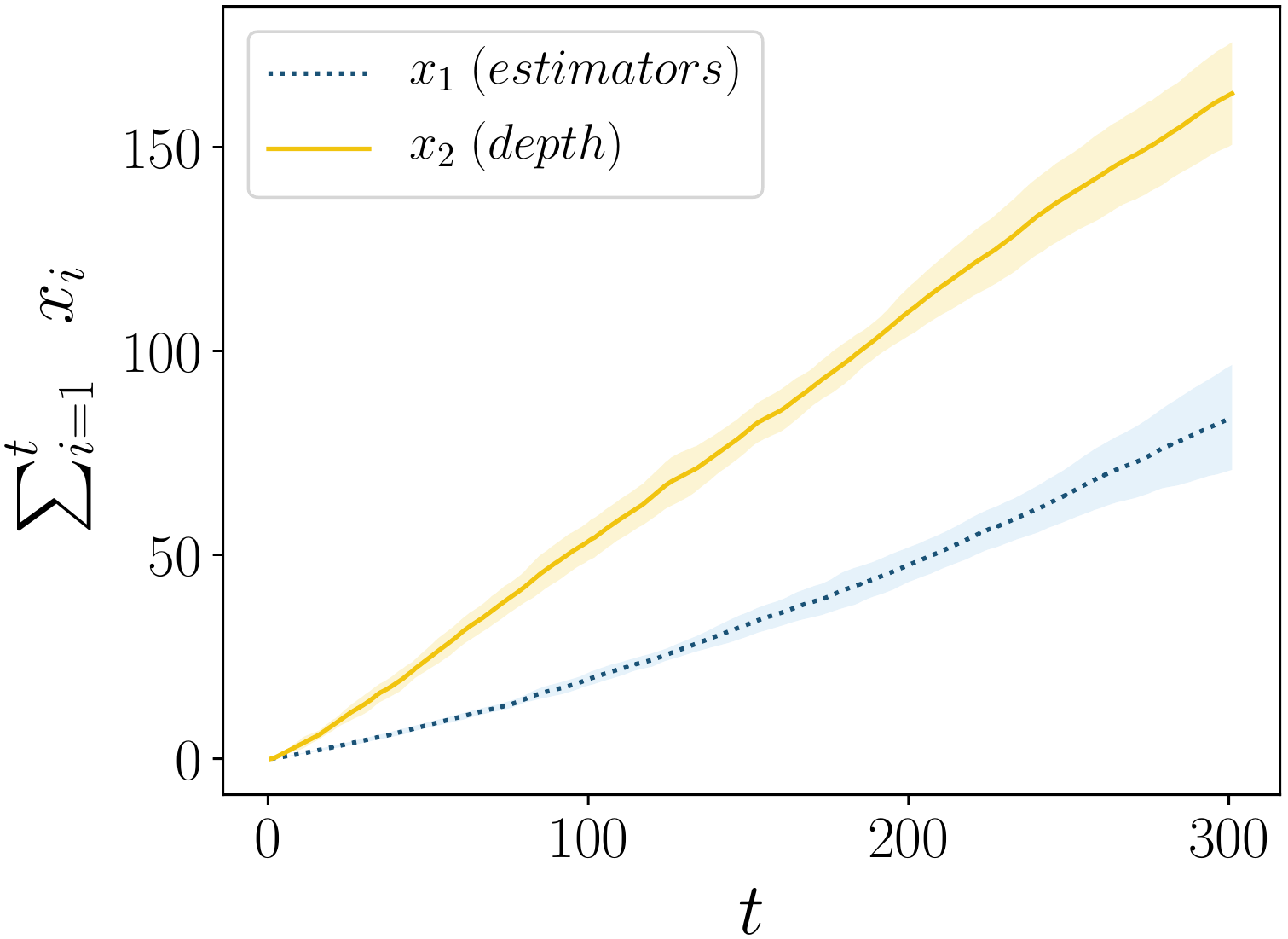}
    \caption{Sum of selected inputs (CA-MOBO)}
    \label{fig:5x3}      
\end{subfigure}
\begin{subfigure}[t]{0.25\linewidth}
    \centering
    \includegraphics[width=1\linewidth]{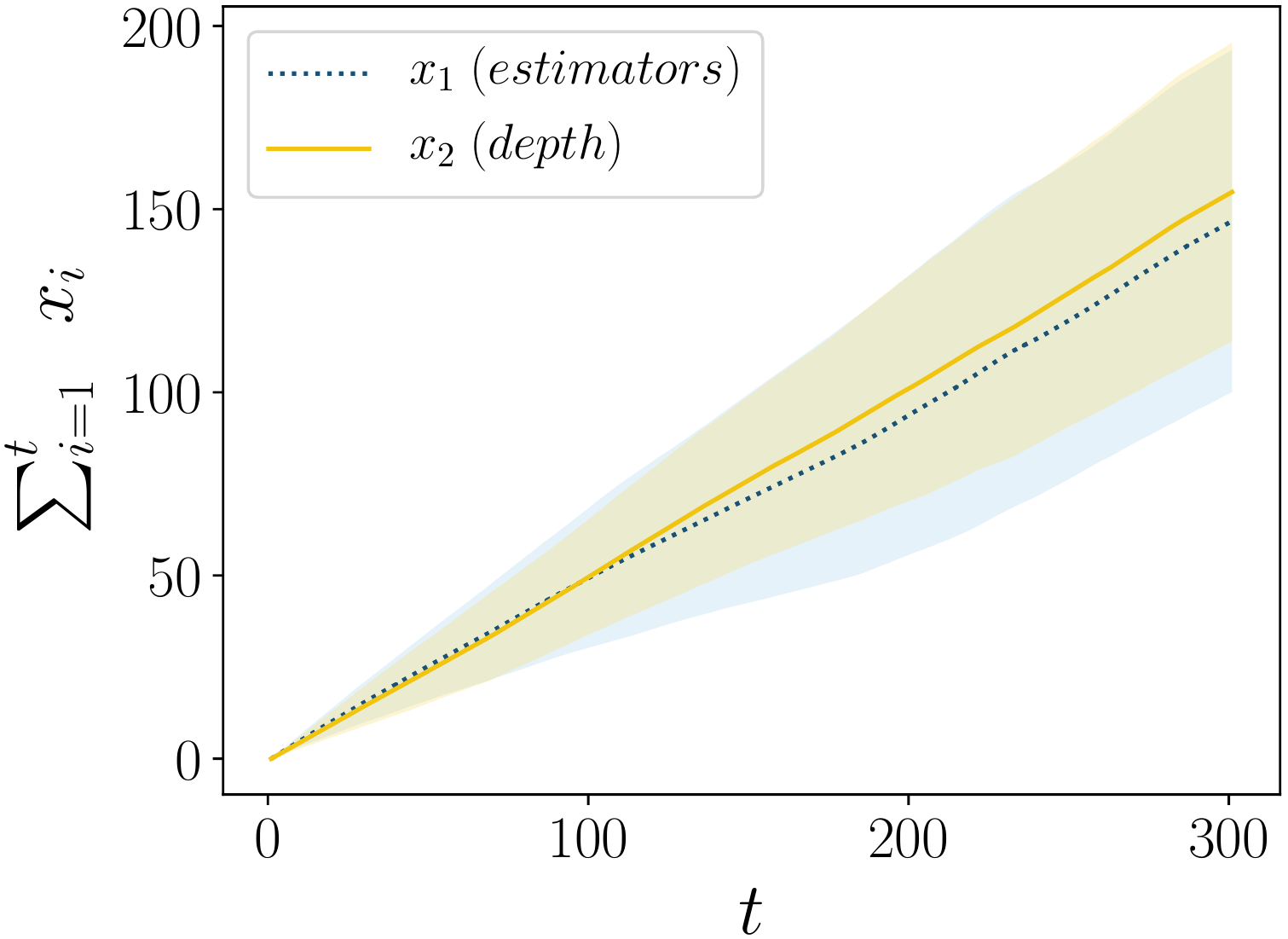}
    \caption{{Sum of selected inputs (MO-UCB)}}
    \label{fig:5x4}
\end{subfigure}
\caption{Optimising a random forest model with two hyperparameters of number of estimators and depth of estimators.}
\label{fig:exp5}
\end{figure*}
\begin{figure*}
\begin{subfigure}[t]{0.26\linewidth}
    \centering
    \includegraphics[width=1\linewidth]{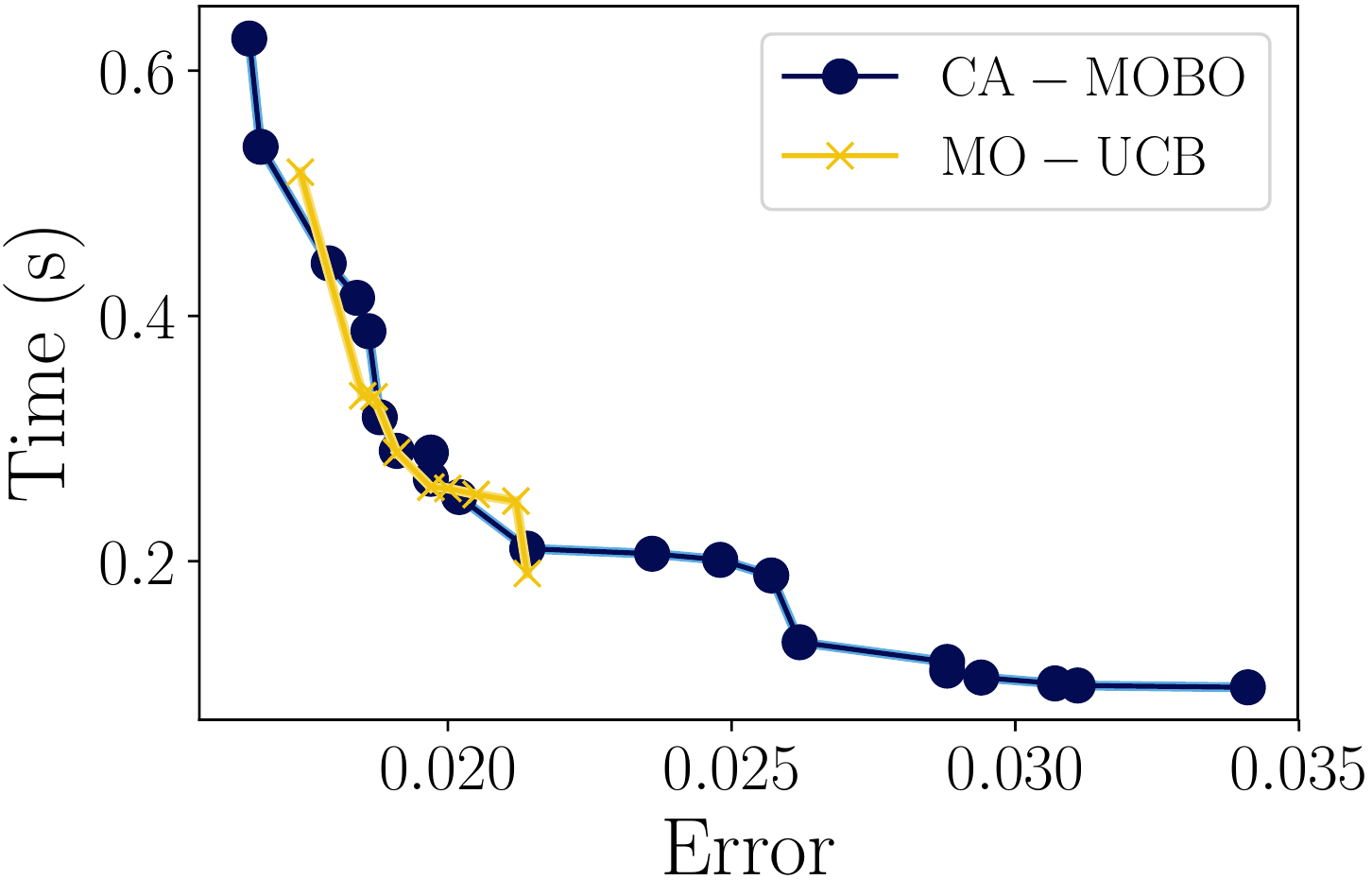}
    \caption{Obtained Pareto front}
    \label{fig:4x1}
\end{subfigure}
\begin{subfigure}[t]{0.25\linewidth}
    \centering
    \includegraphics[width=1\linewidth]{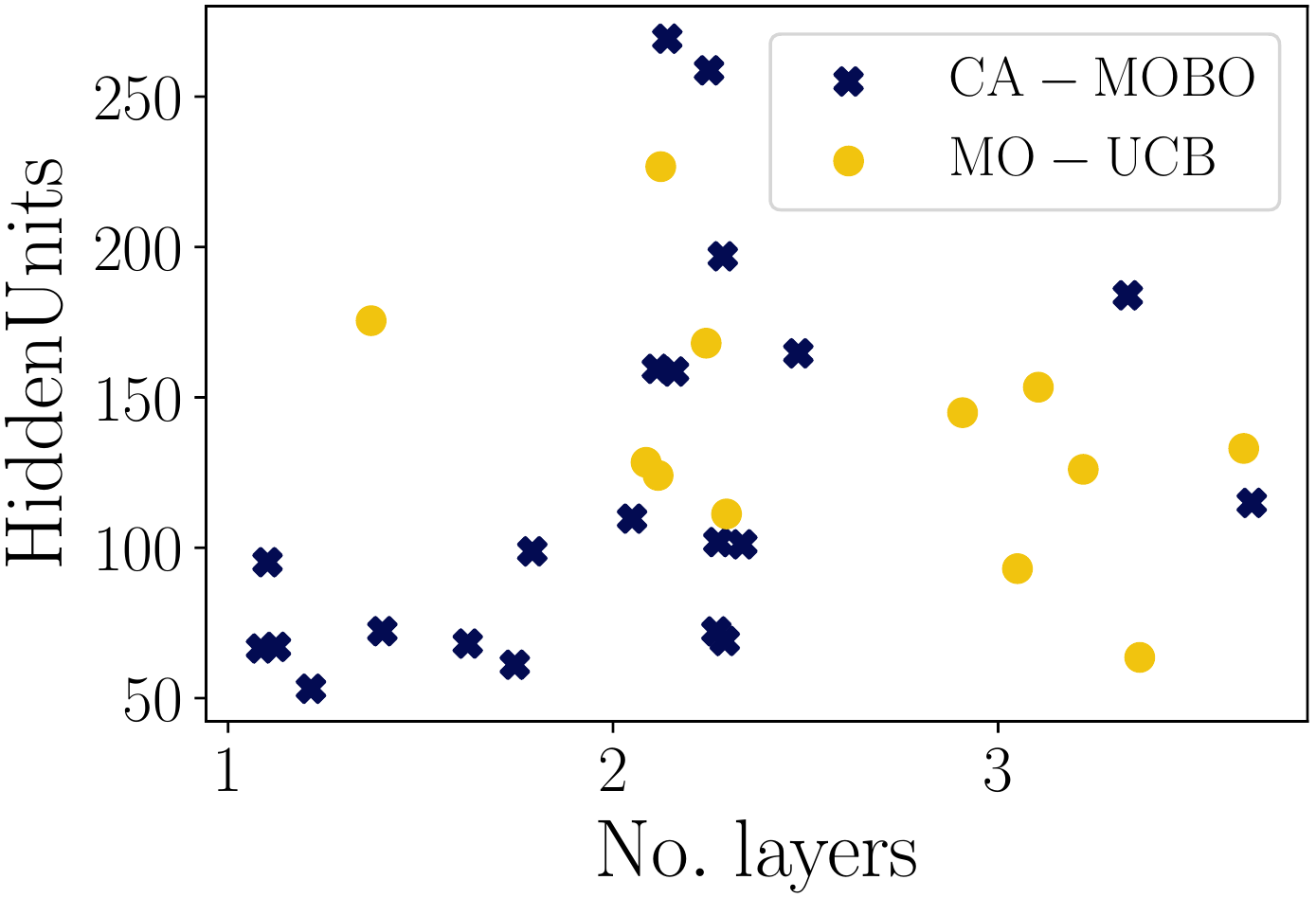}
    \caption{Pareto fronts solutions found in $\mathbb{X}$}
    \label{fig:4x2}       
\end{subfigure}
\begin{subfigure}[t]{0.230\linewidth}
    \centering
    \includegraphics[width=1\linewidth]{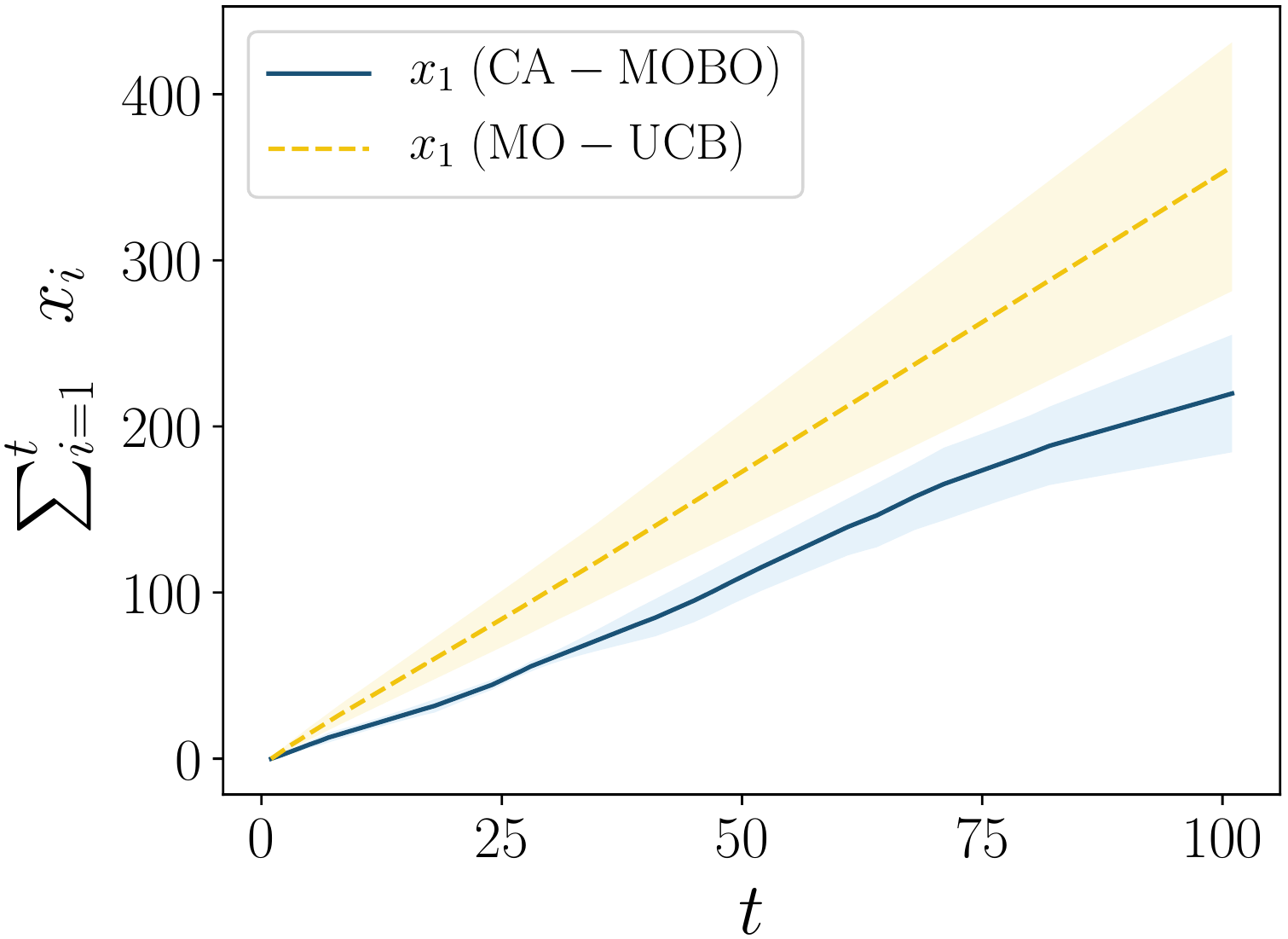}
    \caption{Sum of selected inputs for $x_1$}
    \label{fig:4x3}      
\end{subfigure}
\begin{subfigure}[t]{0.24\linewidth}
    \centering
    \includegraphics[width=1\linewidth]{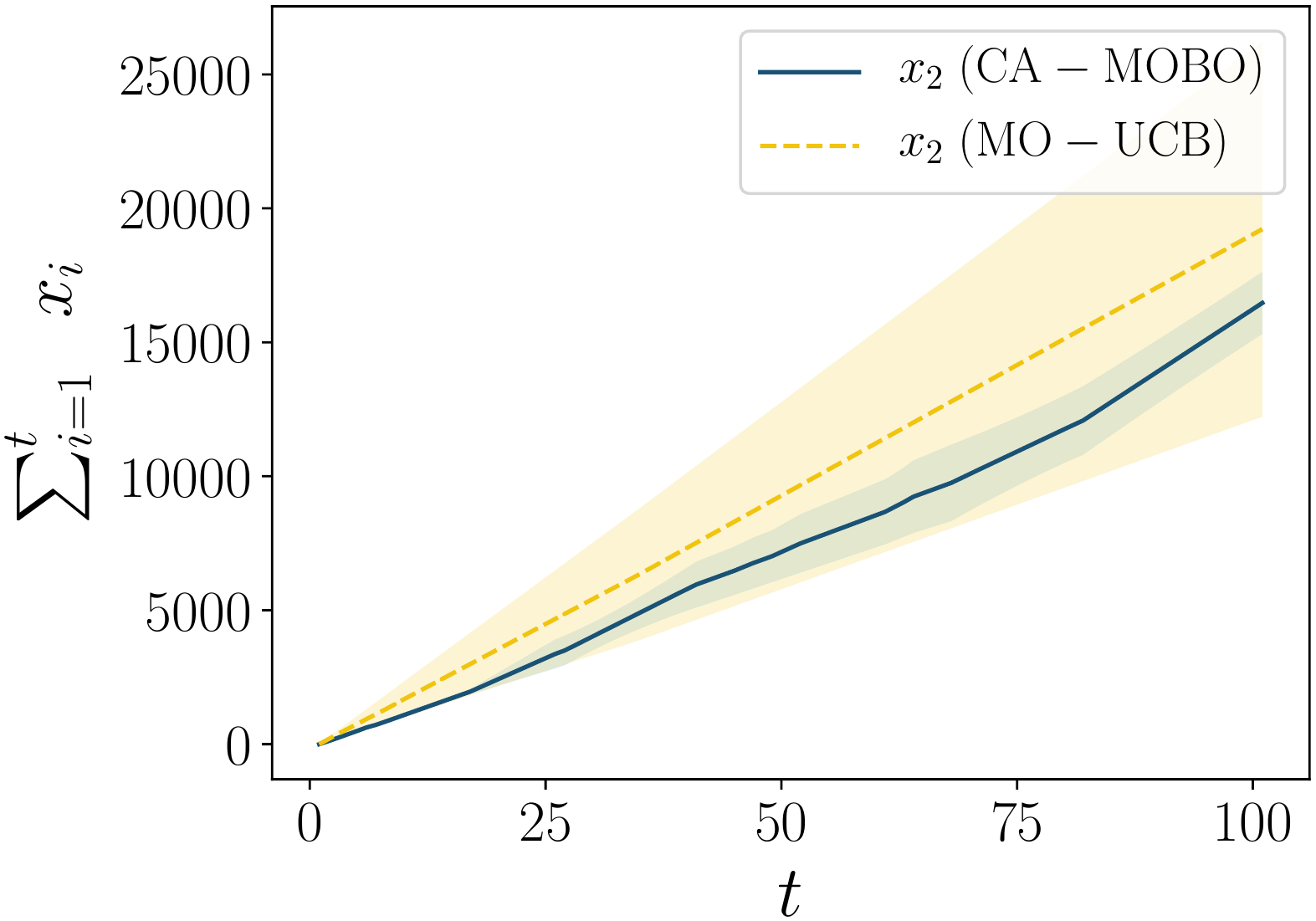}
    \caption{Sum of selected inputs for $x_2$}
    \label{fig:4x4}
\end{subfigure}
\caption{Optimising a neural network with six hyperparameters with two conflicting objectives of prediction error and prediction time.}
\label{fig:exp4}
\end{figure*}
\section{Synthetic Functions}
We first compare performance on minimising synthetic functions. To better illustrate the theoretical analysis on regret in CA-MOBO, we define the average regret at time $t \in \{1,...,T\}$ based on the mean of obtained regrets in $1...t$ iterations, i.e.:
\begin{equation}
\mathcal{R}^{\prime}(t) = \frac{\sum_{t^{\prime}=1}^{t} r(\mathbf{x}_{t^{\prime}},{\bm \theta}_{t^{\prime}})}{t}
\label{eq:avgR}
\end{equation}
The average regret is easier to interpret than the instantaneous regret. 
\indent{} \par
We start our experiments with Zitzler-Deb-Thiele's function N. 3 \cite{zitzler2000comparison} with $5-$dimensional input and $2-$dimensional output. Without loss of generality, we define the cost-aware preferences as $\mathcal{I} = (1,2,3,4,5)$ - i.e. selecting $x_1$ value as a input from dimension $1$ of the search space is more expensive than selecting the same value of $x$ from dimension $2$ and so on. Figure \ref{fig:1x1} shows the whole Pareto front as the ground-truth. Figure \ref{fig:1x5} and \ref{fig:1x7} show that both CA-MOBO and MO-UCB obtain regions of Pareto front, and both methods achieved approximately the same dominated hypervolume  but with different exploration strategies on the search space (see Figure \ref{fig:1x2}). Figure \ref{fig:1x6} shows the value of $\sum_{t=1}^{T}\ x_i,\ i \in \{1,...,5\}$ with respect to $t$ and it confirms that CA-MOBO is complying with the cost-aware constraints since the cumulative selected values of $x_1$ (the most expensive dimension of search space) is $\sum_{t=1}^{T=500}\ x_1 \approx 17.8$ for CA-MOBO while in Figure \ref{fig:1x8} this values is approximately $\sum_{t=1}^{T=500}\ x_1 \approx 85.2$ for MO-UCB. Moreover, based on Figure \ref{fig:1x6}, our proposed method successfully followed the ordering of dimensions of search space based on cost-aware constraint $\mathcal{I}$ as the highest usage is for dimension $5$ and the lowest one is for dimension $1$. Whereas in Figure \ref{fig:1x8} with no cost-awareness, $x_3$ has the highest value of cumulative usage during the optimisation. It is noteworthy to mention that while both methods approximately achieved the same value of dominated hypervolume in finding Pareto front, given the cost-aware constraints, CA-MOBO uses smaller amounts of expensive inputs during the optimisation. Figure \ref{fig:1x3} shows the average regret as defined in (\ref{eq:avgR}). MO-UCB achieves a better regret initially due to its capacity for exploration in more expensive regions of search space that allows it to find higher number of dominating solutions. CA-MOBO obtains a better regret at the end of the optimisation due to its compliance with the cost-aware constraints and gradually explores more expensive regions.
Likewise, Figure \ref{fig:1x4} illustrates the cumulative regret incurred by CA-MOBO comparing to MO-UCB.
\indent{} \par
Figure \ref{fig:exp3} shows the results for minimising Cross-in-tray function as the first objective and Hölder table function \cite{jamil2013literature} as the second objective:
{\small
\[
f_1(\mathbf x) = -10^{-4}\Big[\Big|\mathrm{sin}(x_1)\mathrm{sin}(x_2)\mathrm{exp}(\big|100-\frac{\sqrt{x_1^2+x_2^2}}{\pi}\big|)  \Big|+1\Big]^{0.1}
\]
\[
f_2(\mathbf x) = -\Big|\mathrm{sin}(x_1)\mathrm{cos}(x_2)\mathrm{exp}\big(\Big|1-\frac{\sqrt{x_1^2+x_2^2}}{\pi}\Big|\big)  \Big|
\]
}
For this problem, we define $\mathcal{I} = \{1,2\}$ as the cost-aware constraint. Unlike the previous experiment, this constraint results in better initial observations for CA-MOBO as Figure \ref{fig:3x2} illustrates the initial values of dominated hypervolume for CA-MOBO is higher than MO-UCB. Correspondingly, Figure \ref{fig:3x3} and Figure \ref{fig:3x4} show an advantage in average regret and cumulative regret for CA-MOBO at initial iterations of optimisation. Comparing Figure \ref{fig:3x6} and Figure \ref{fig:3x8}, $\sum_{t=1}^{T=500}\ x_1 < \sum_{t=1}^{T=500}\ x_2$ for CA-MOBO (with low uncertainty) and $\sum_{t=1}^{T=500}\ x_1 \approx \sum_{t=1}^{T=500}\ x_2$ for MO-UCB. That implies CA-MOBO complies with cost-aware constraints while achieving the same percentage of dominated hypervolume in comparison to MO-UCB with no cost-awareness.
 
\subsection{Hyperparameters of Random Forest}
For random forest, we have defined two objectives - training time and prediction error. The hyperparameters are the number of estimators ($x_1 \in [1,100]$) and the depth of estimators ($x_2 \in [1,100]$). The defined estimators are decision trees. The Scikit-Learn python package \cite{pedregosa2011scikit} is used for implementation.\indent{} \par
We define the cost-aware constraints for CA-MOBO as $\mathcal{I} = \{1,2\}$ - i.e. the changes in number of estimators are considered to be more expensive compared to the depth of estimators. Figure \ref{fig:5x1} compares the Pareto fronts found by CA-MOBO and MO-UCB. Our proposed method achieved more diverse Pareto front in comparison to MO-UCB, specifically when the training time is low and error high. We believe the reason for more diverse solutions in the region with low training time and high error is related to the cost-aware constraints on the search space which in turn discourages the optimiser to initially explore the regions of the search space with high values of estimators for the random forest. As a result, initial solutions with high error are favored for CA-MOBO. However, MO-UCB obtains more number of Pareto front solutions in the region with low error after $t=300$ iterations.  Figure \ref{fig:5x2} confirms our conclusion as most of the selected solutions by CA-MOBO in $\mathbb{X}$ (search space) have low values in number of estimators ($x_1$). Figure \ref{fig:5x3} shows the sum of all the selected inputs from the search space - CA-MOBO has  $\sum_{i=1}^{T}\ x_1 \approx 71.4$ the sum over the number of estimators, which increases almost to twice the value for MO-UCB  ($\sum_{i=1}^{T}\ x_1 \approx 141.8$)(see Figure \ref{fig:5x4}).

\subsection{Fast and Accurate Neural Network}
In neural networks, low prediction error and small prediction time is desirable, however, these are conflicting objectives.
We define the aim of this experiment to find fast and accurate neural networks by tuning six hyperparameters on the MNIST dataset \cite{lecun1998gradient}.  This problem first proposed by \cite{hernandez2016predictive}  and we have slightly modified this experiment to fit in our problem framework. The hyperparameters to be tuned are:  Number of hidden layers ($x_1 \in [1,4]$), the number of hidden units per layer ($x_2 \in [50,300]$), the learning rate ($x_3 \in (0,0.2]$), amount of dropout ($x_4 \in [0.4,0.8]$), and the level of $l_1\ (x_5 \in (0,0.1])$ and $l_2\ (x_6 \in (0,0.1])$ regularization. 
We consider a feed-forward  networks with ReLUs at the hidden layers and a soft-max output layer. The networks are coded in the Keras library and trained using Adam \cite{kingma2014adam} with a batch size of $4000$ instances in $64$ epochs. 
\indent{}\par
We define the cost-aware constraints for CA-MOBO as $\mathcal{I} = \{1,2\}$, which means dimension $1$ from the search space is more expensive comparing to second dimension - i.e. increasing the number of hidden layers is more expensive than the number of hidden units per layer. Figure \ref{fig:4x1} illustrates more diverse solutions for CA-MOBO as compared to MO-UCB, mainly in the region with high error and low test time due to the constraints imposed by cost-awareness initially. As in the random forest experiment, Figure \ref{fig:4x2} shows our proposed method favors cheaper regions of $\mathbb{X}$. Finally, Figure \ref{fig:4x3} indicates in $300$ iterations of the optimisation for CA-MOBO with $\sum_{i=1}^{T}\ x_1 \approx 205.7$, whereas for MO-UCB $\sum_{i=1}^{T}\ x_1 \approx 349.5$. Figure \ref{fig:4x4} illustrates the sum of inputs over number of hidden units ($x_2$) for both algorithms.
\section{Conclusion}
We have proposed  a  novel  algorithm  for  multi-objective Bayesian optimization to incorporate non-uniform black-box function evaluations. 
We have defined cost-aware constraints over the search space in order to 
model the varied cost of function evaluations for different combination of inputs from the search space. CA-MOBO initially explores less expensive regions of the search space and gradually moves towards the more expensive regions of the search space. Experimental results 
show the effectiveness of our algorithm.
\bibliographystyle{aaai}
\bibliography{Main}
\section{Supplementary Materials}
\subsection{Proof of Theorem 1}
Considering $\alpha({\bf x},{\bm \theta}_t,t)$ as defined in (9) (CA-MOBO acquisition function), $\mathbb{E}[\mathcal{R}(T)]$ can be written as:
\begin{align*}
\mathbb{E} \Big[
\sum_{t=1}^{T} \Big(\mathrm{\maxx_{{\bf x} \in \mathcal{X}}}\ \  S_{{\bm \theta}_t} \big(f({\bf x})\big)\big(1- C({\bf x},t)\big)-S_{{\bm \theta}_t} \big(f({\bf x}_t)\big)...\\
...\big(1-C({\bf x_t},t)\big)\Big) \\
\leq  
\underbrace{
\mathbb{E} \Bigg[
\sum_{t=1}^{T} \Big(  
\alpha({\bf x}_t,{\bm \theta}_t,t) - S_{{\bm \theta}_t} (f({\bf x}_t))\Big).\Big(1-C({\bf x}_t,t)\Big)
 \Bigg]}_{\mathrm{(I)}}+ \\
 \underbrace{
 \mathbb{E} \Bigg[
\sum_{t=1}^{T} \Big(  
S_{{\bm \theta}_t} (f({\bf x}^*_t)) - \alpha({\bf x}^*_t,{\bm \theta}_t,t)\Big).\Big(1-C({\bf x}^*_t,t)\Big)
 \Bigg]}_{\mathrm{(II)}}
\end{align*}
given that:
$
{\bf x}_t = \argmax_{{\bf x} \in \mathbb{X}}\  \alpha({\bf x},{\bm \theta}_t,t)
$ and 
$
{\bf x}^*_t = \argmax_{{\bf x} \in \mathbb{X}}\ \ S_{{\bm \theta}_t} (f({\bf x}))\big(1- C({\bf x},t)\big).
$
Based on the definition of ${\bf x}_t$ and ${\bf x}^*_t$, it is clear that $\alpha({\bf x}_t,{\bm \theta}_t,t) \geq \alpha({\bf x}^*_t,{\bm \theta},t)$ since $\bf x_t$ results in the highest value of $\alpha({\bf x}_t,{\bm \theta}_t,t)$. Based on $\mathrm{(I)}$ and $\mathrm{(II)}$, we are defining corollary \ref{cor:Sup1} and corollary \ref{cor:Sup2}.
\begin{cor_sup}
It can be proved that $\mathrm{(II)}$ is bounded as:
\begin{align*}
 \mathbb{E} \Bigg[
\sum_{t=1}^{T} \Big(  
S_{{\bm \theta}_t} (f({\bf x}^*_t)) - \alpha({\bf x}^*_t,{\bm \theta}_t,t)\Big).\Big(1-C({\bf x}^*_t,t)\Big)
 \Bigg]...\\
...\leq \frac{\pi^2}{6} \mathbb{E}[{U}_{\bm \theta}] M
\end{align*}
\label{cor:Sup1}
where $U_{\bm \theta}$ is an arbitrary distribution over the weights of Chebyshev scalarisation.
\end{cor_sup}

\begin{figure*}
\begin{subfigure}[t]{0.255\linewidth}
    \centering
    \includegraphics[scale=0.314]{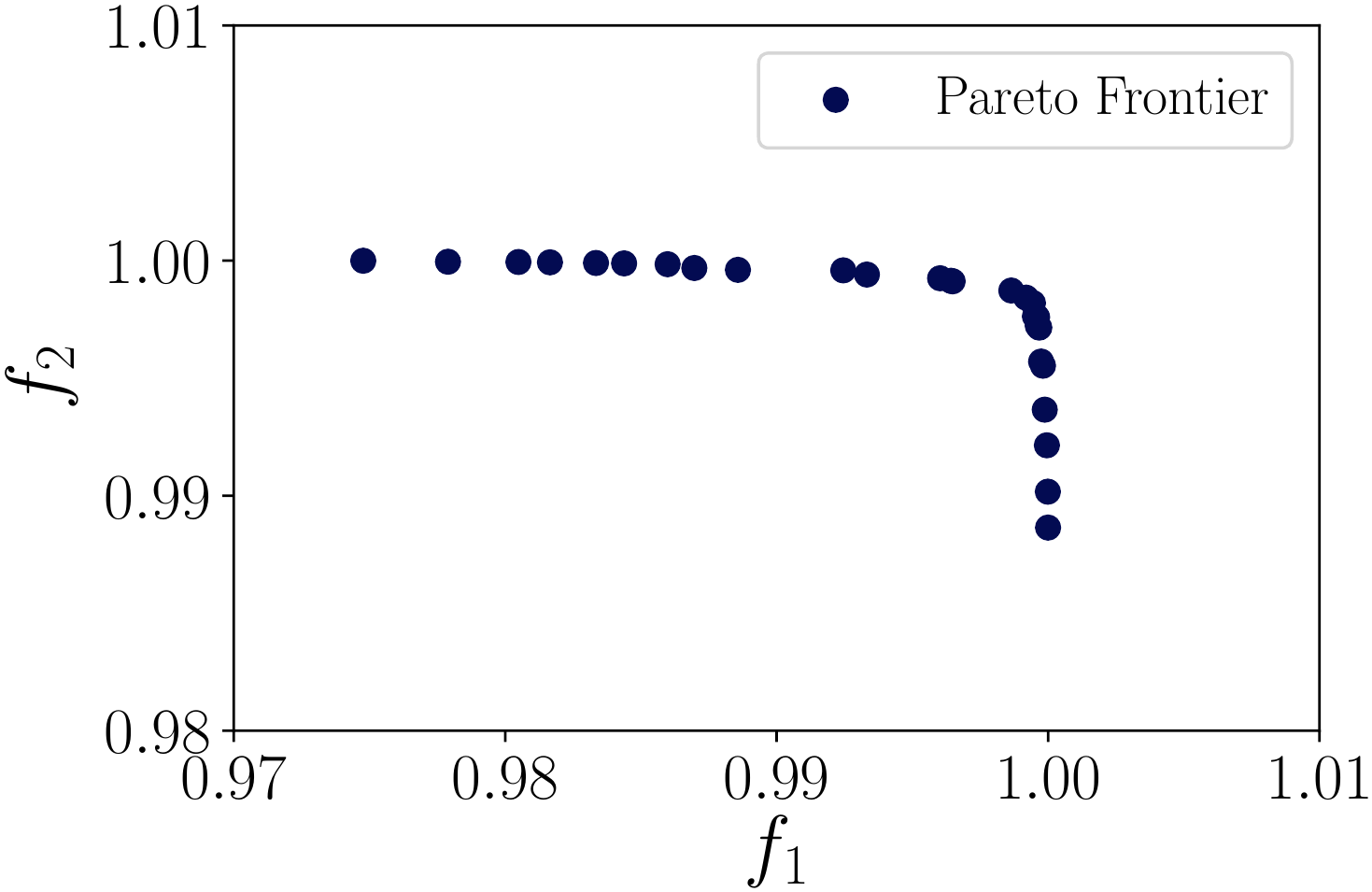}
    \caption{Full Pareto front}
    \label{fig:2x1}
\end{subfigure}
\begin{subfigure}[t]{0.245\linewidth}
    \centering
    \includegraphics[width=1\linewidth]{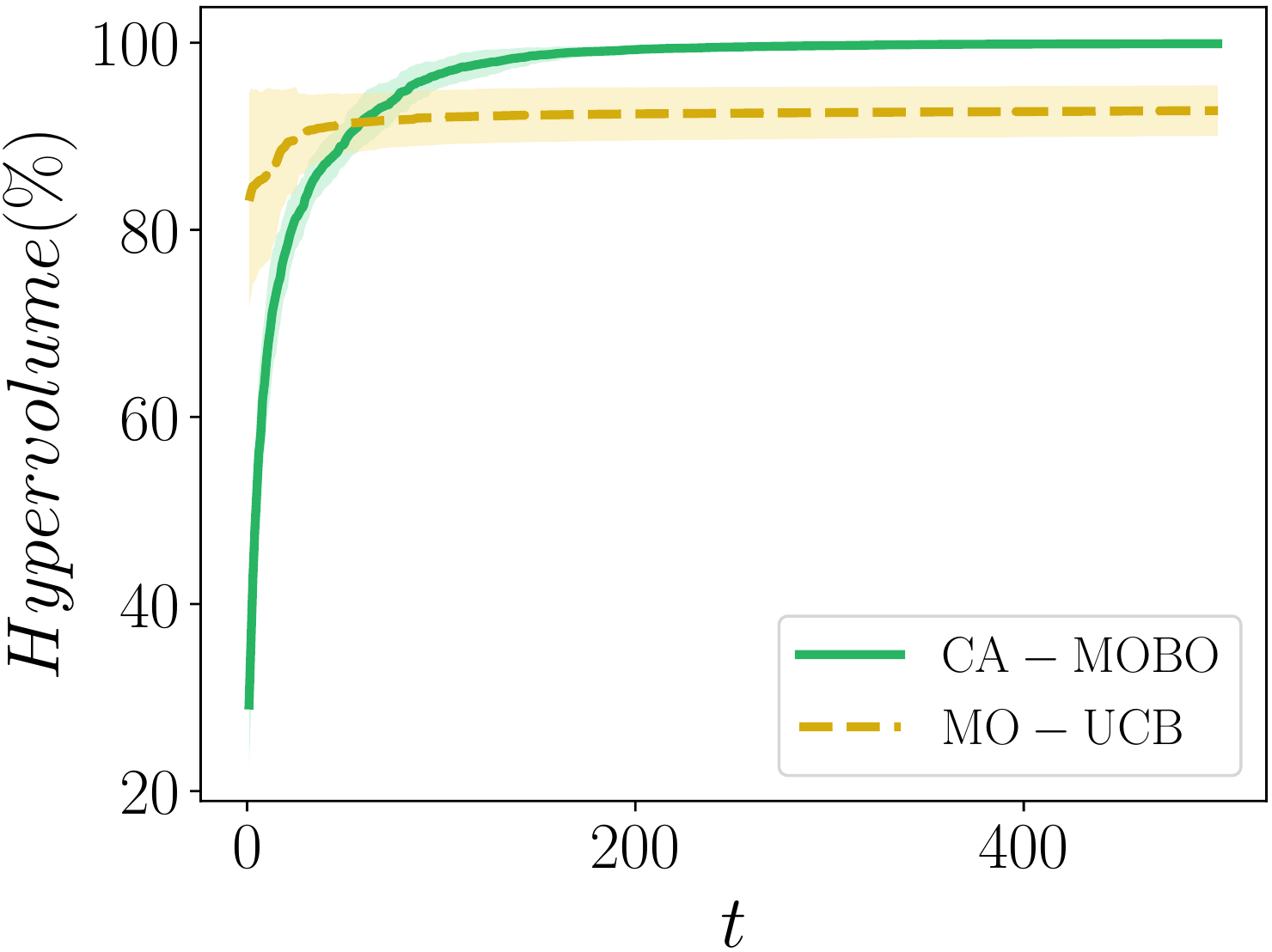}
    \caption{Dominated Hypervolume}
    \label{fig:2x2}
\end{subfigure}
\begin{subfigure}[t]{0.245\linewidth}
    \centering
    \includegraphics[width=1\linewidth]{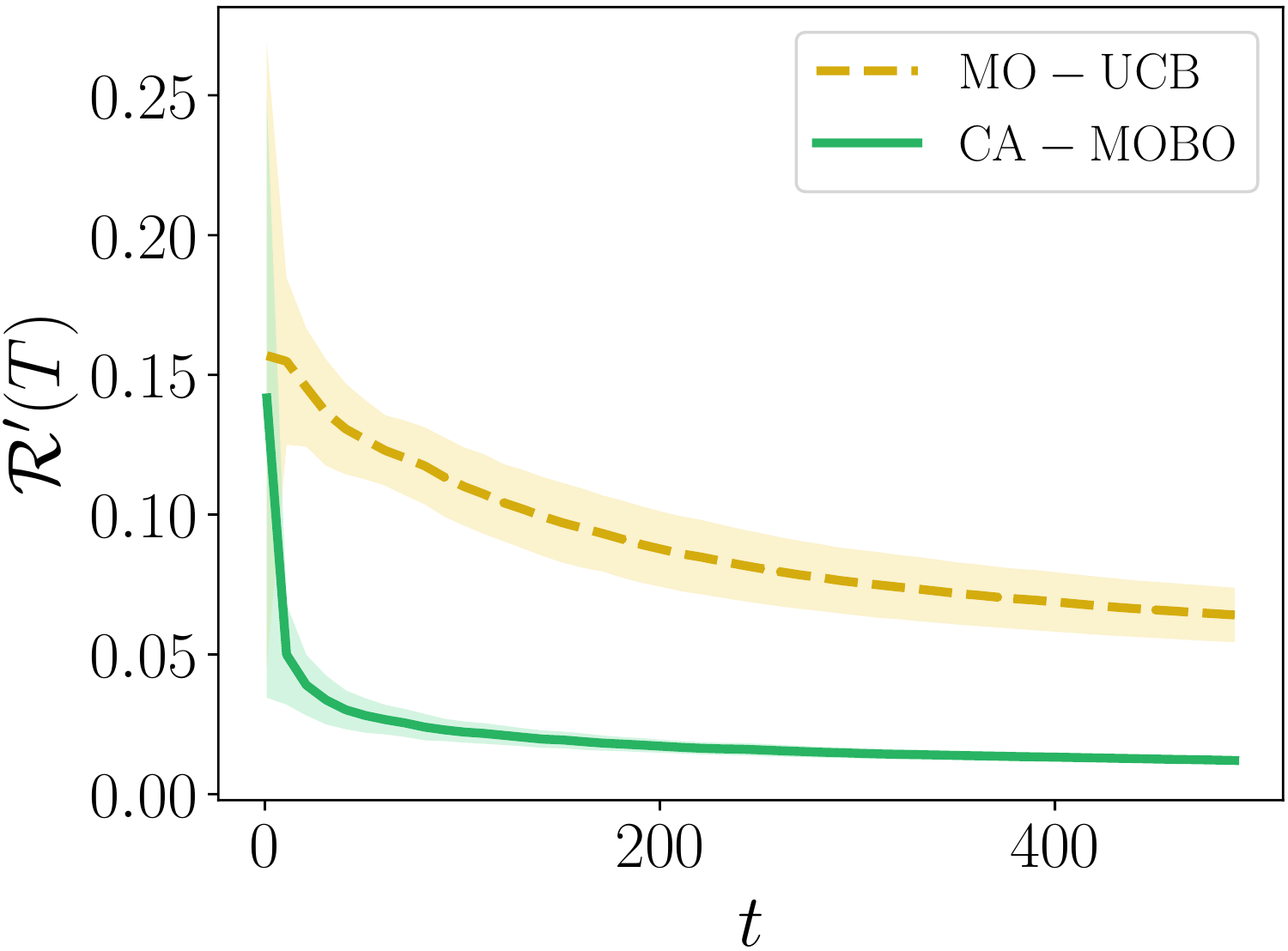}
    \caption{Average Regret}
    \label{fig:2x3}      
\end{subfigure}
\begin{subfigure}[t]{0.241\linewidth}
    \centering
    \includegraphics[width=1\linewidth]{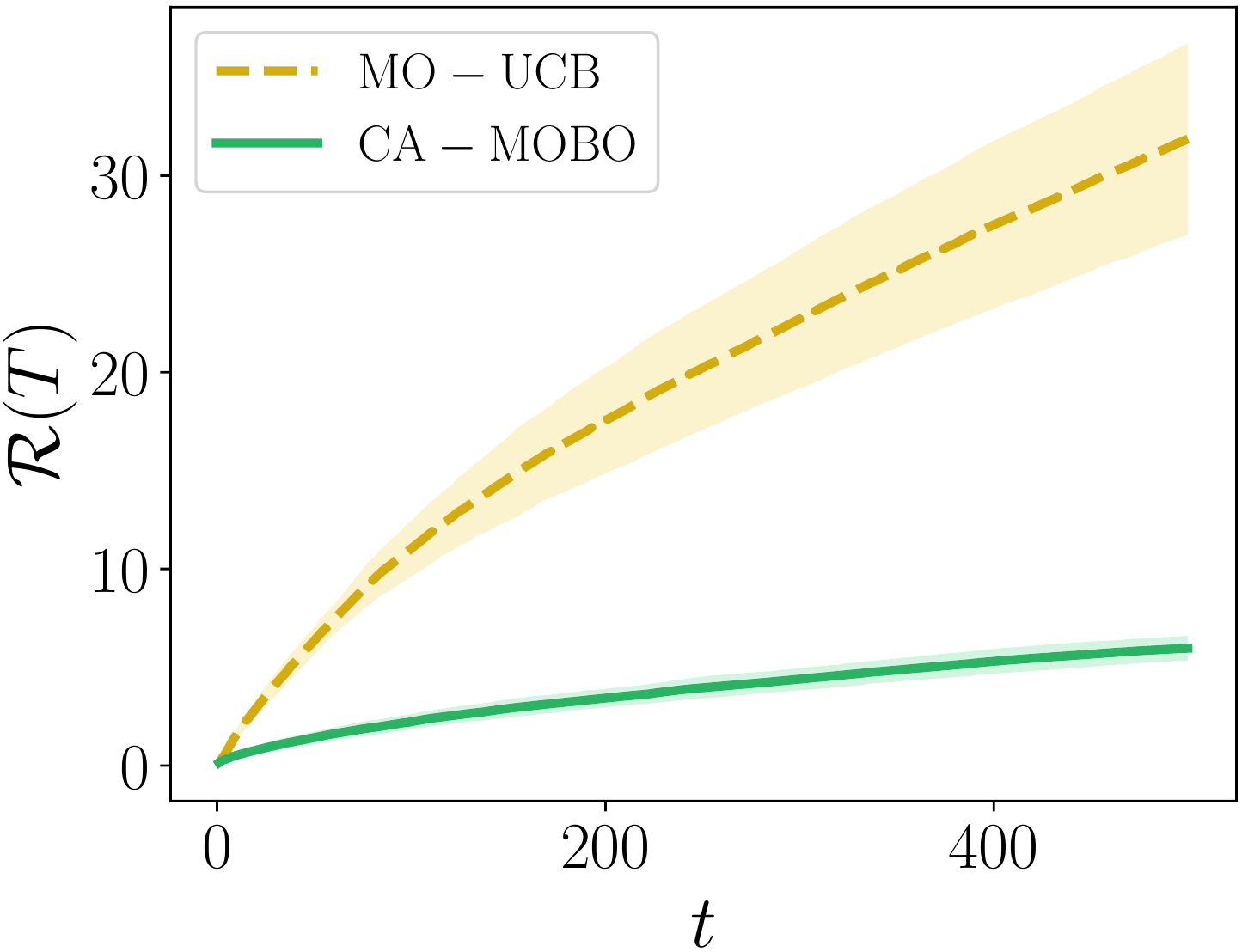}
    \caption{Cumulative Regret}
    \label{fig:2x4}
\end{subfigure}
\begin{subfigure}[t]{0.245\linewidth}
    \centering
    \includegraphics[width=1\linewidth]{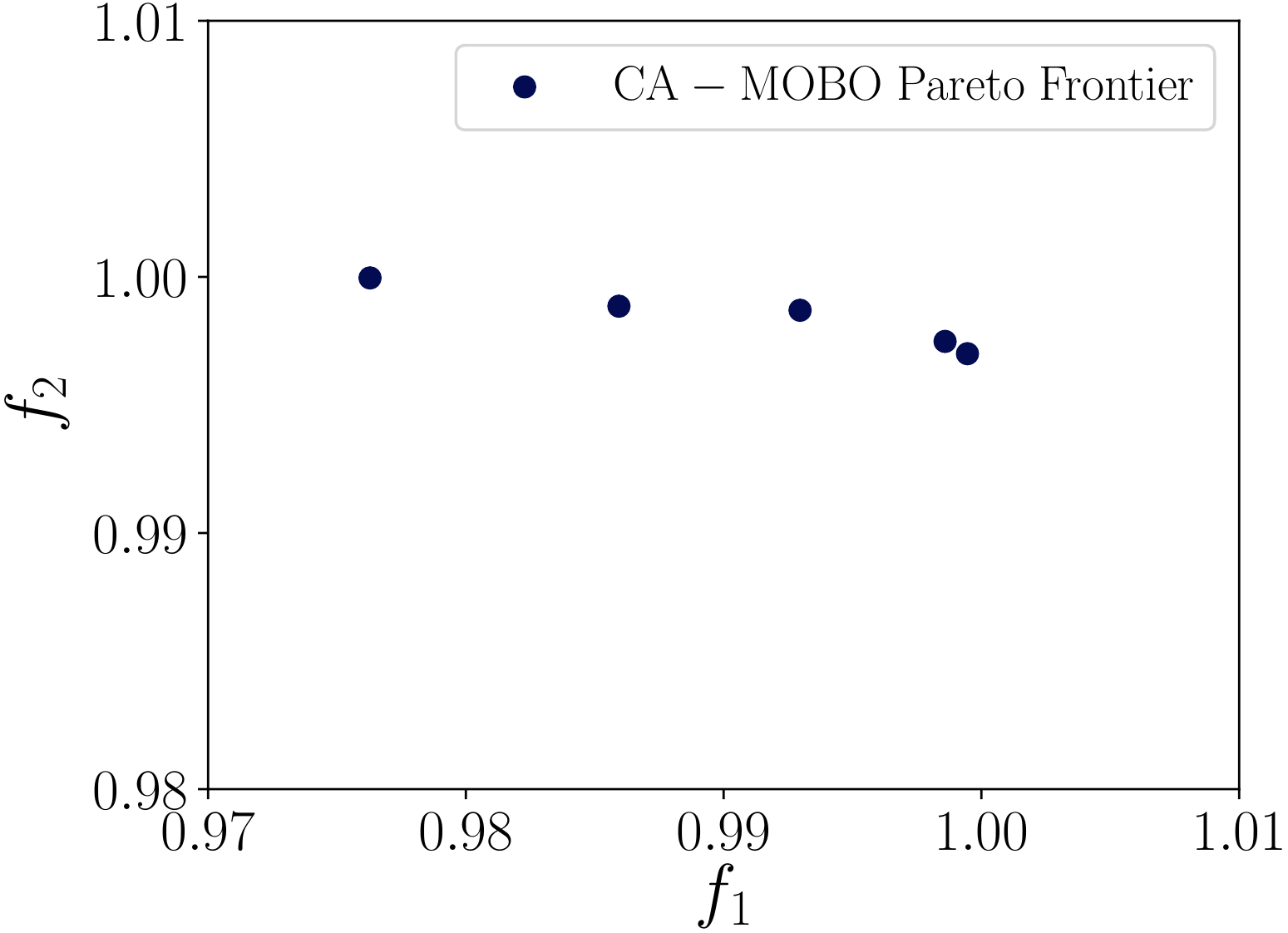}
    \caption{Pareto front (CA-MOBO) }
    \label{fig:2x5}       
\end{subfigure}
\begin{subfigure}[t]{0.245\linewidth}
    \centering
    \includegraphics[width=1\linewidth]{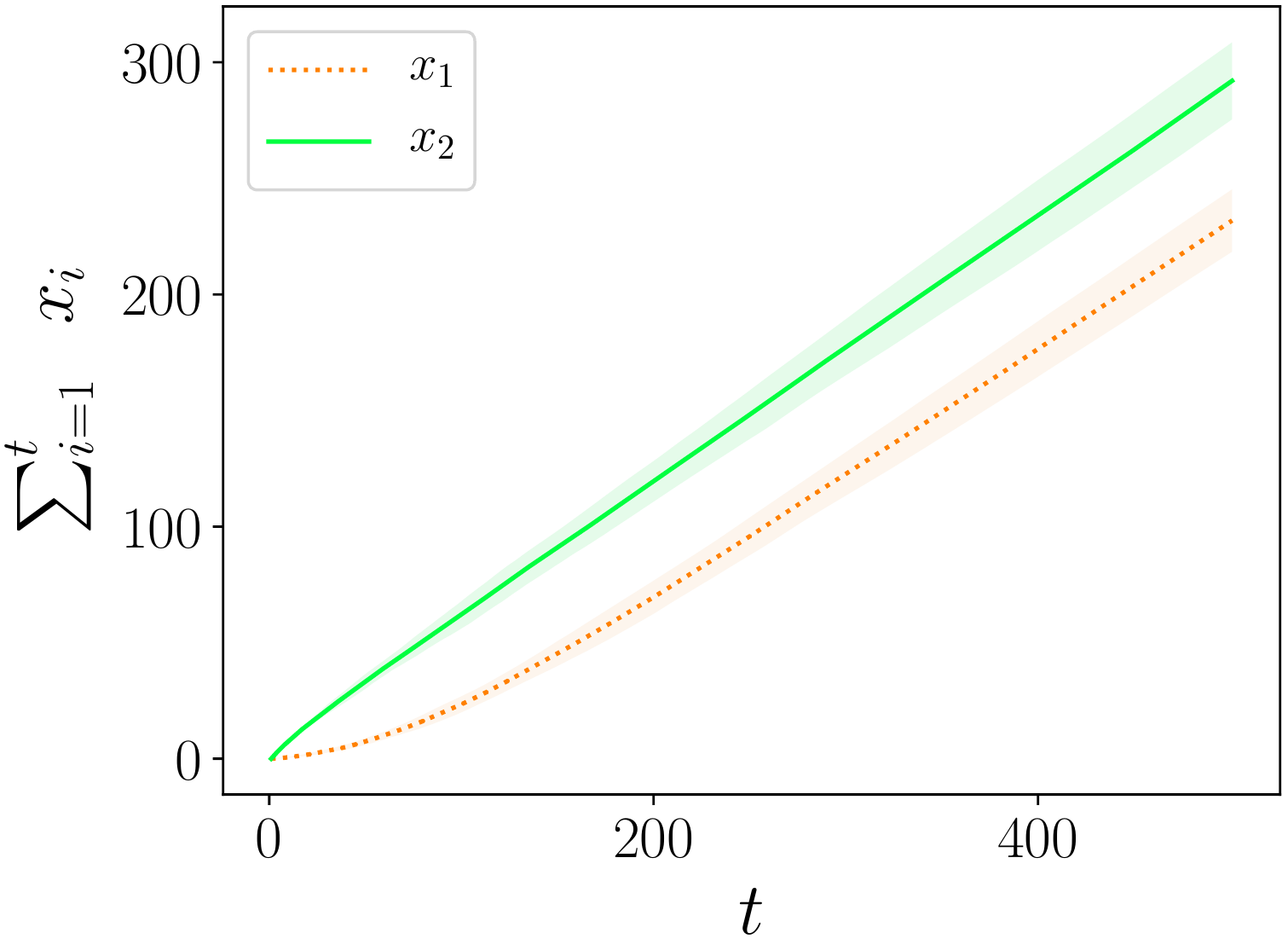}
    \caption{Sum of selected inputs (CA-MOBO)}
    \label{fig:2x6}      
\end{subfigure}
\begin{subfigure}[t]{0.245\linewidth}
    \centering
    \includegraphics[width=1\linewidth]{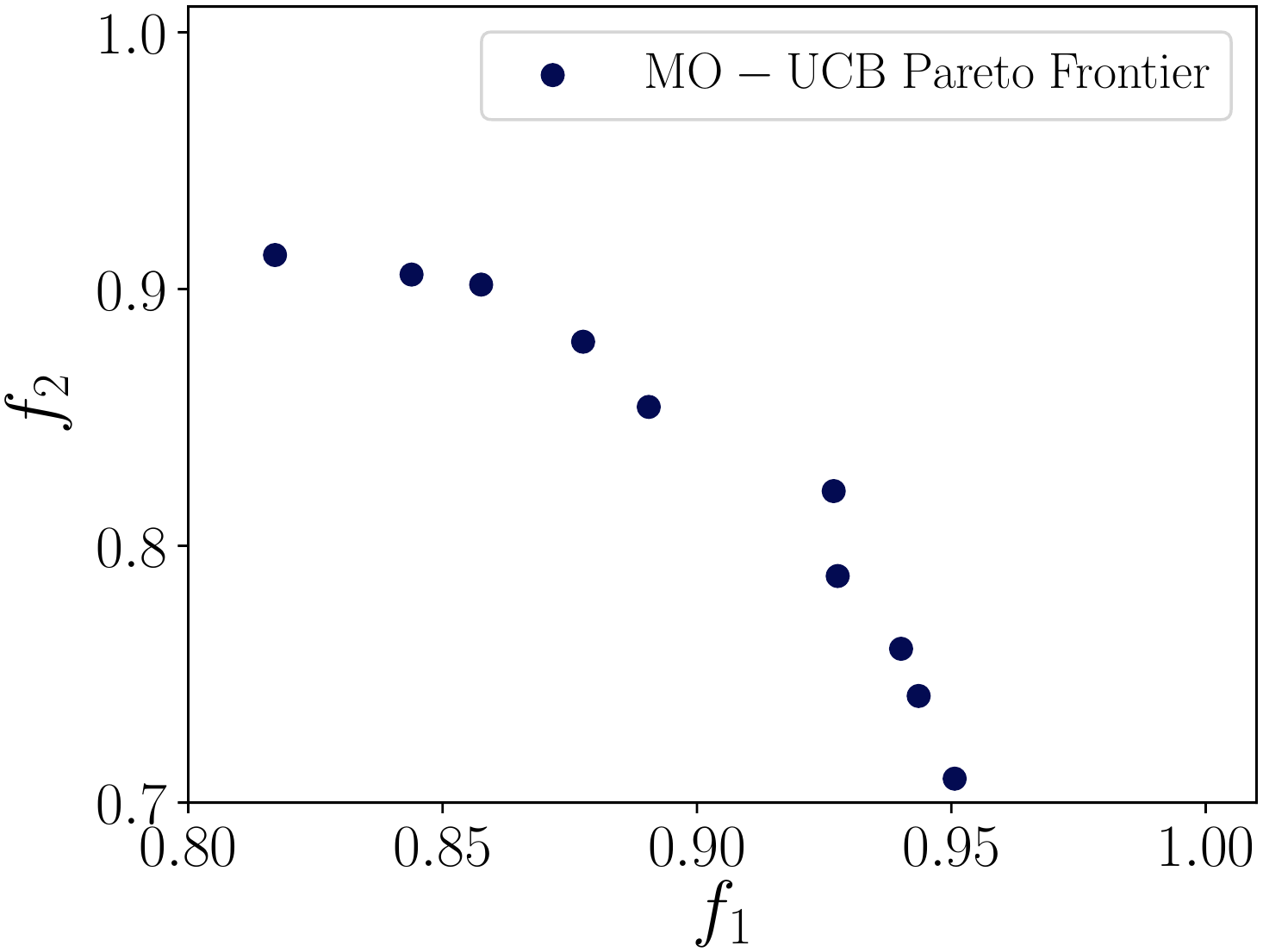}
    \caption{Pareto front (MO-UCB)}
    \label{fig:2x7}
\end{subfigure}
\begin{subfigure}[t]{0.25\linewidth}
    \centering
    \includegraphics[width=1\linewidth]{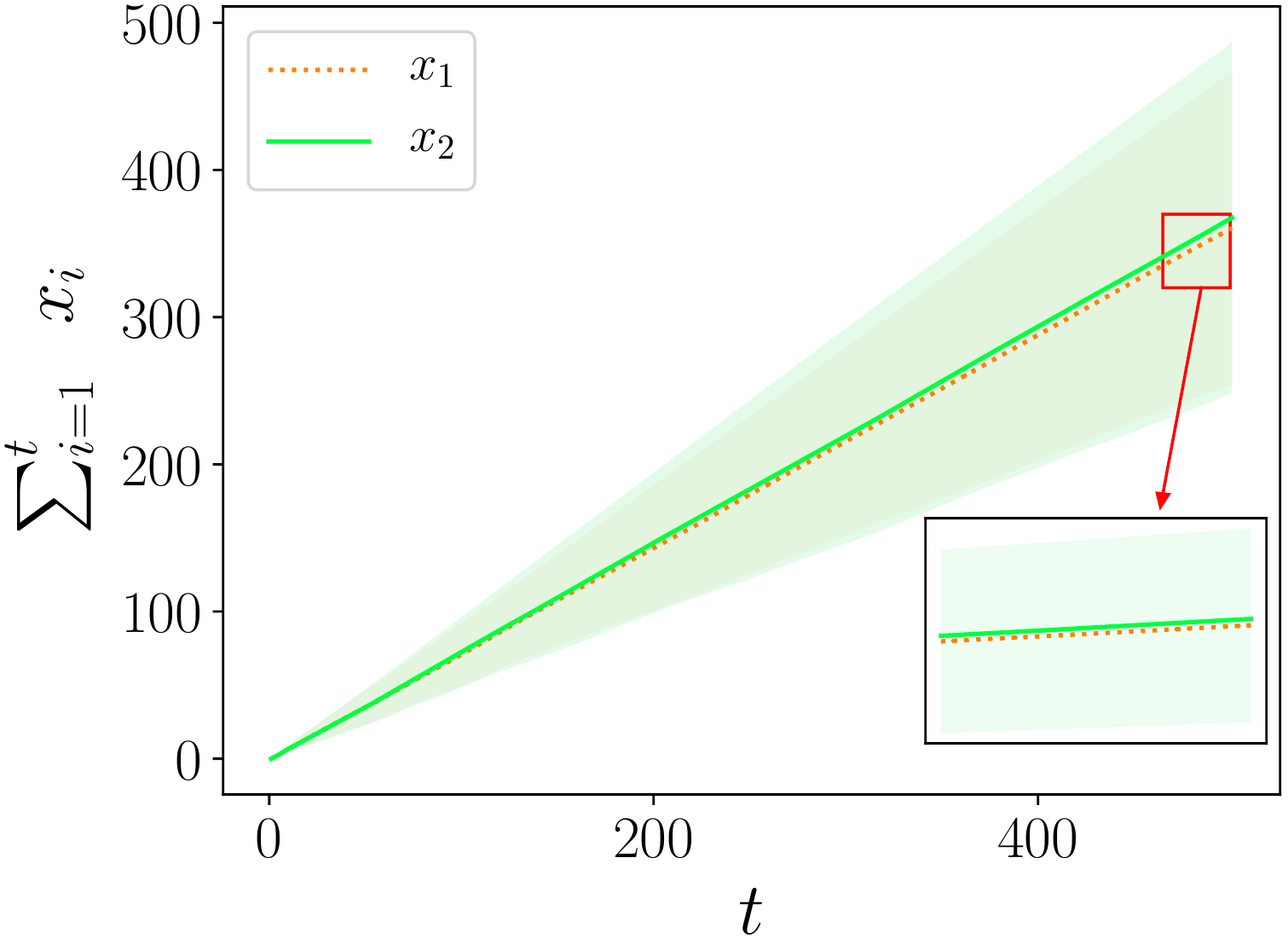}
    \caption{Sum of selected inputs (MO-UCB)}
    \label{fig:2x8}      
\end{subfigure}
\caption{Minimising Matyas function  as first objective and Booth function as the second objective. Figure \ref{fig:2x1} shows the full Pareto front for this problem. Figure \ref{fig:2x2} show the comparison of dominated hypervolume. Figure \ref{fig:2x3} and Figure \ref{fig:2x4} demonstrate the average and cumulative regret respectively. Figure \ref{fig:2x5} and \ref{fig:2x7} shows the obtained Pareto front for CA-MOBO and MO-UCB respectively. Figure \ref{fig:2x5} confirms the role of cost-aware constraints as $\sum_{t=1}^{T} x_1 < \sum_{t=1}^{T} x_2,\ \forall t$ with low uncertainty. Whereas Figure \ref{fig:2x8} indicates higher usage of dimension $1$ during the optimisation for MO-UCB. }
\label{fig:exp2}
\end{figure*}
\begin{proof}
We start the proof by:
\begin{align*}
\mathbb{E} \Big[
\big(  
S_{{\bm \theta}_t} (f({\bf x}^*_t)) - \alpha({\bf x}^*_t,{\bm \theta}_t,t)\big).(1-C({\bf x}^*_t,t))
 \Big]\\ 
\leq 
 \mathbb{E} \Big[
\big(  
S_{{\bm \theta}_t} (f({\bf x}^*_t)) - \alpha({\bf x}^*_t,{\bm \theta}_t,t)\big).(1-C({\bf x}^*_t,t))
 \Big]_{+}\\
\leq 
 \mathbb{E}   \Big[ \sum_{{\bf x} \in \mathbb{X}}
\big(  
S_{{\bm \theta}_t} (f({\bf x}^*_t)) - \alpha({\bf x}^*_t,{\bm \theta}_t,t)\big).(1-C({\bf x}^*_t,t))
 \Big]_{+} 
\end{align*}
where $[z]_+ = \mathrm{max}(0,z)$. Based on the assumption of $\mathbb{E}[U_{\bm \theta}]$-Lipschitz for $S$ scalarisation function, it can be concluded that:
\begin{align*}
\mathbb{E}   \Big[ \sum_{{\bf x} \in \mathcal{X}}
\big(  
S_{{\bm \theta}_t} (f({\bf x}^*_t)) - \alpha({\bf x}^*_t,{\bm \theta}_t,t)\big).(1-C({\bf x}^*_t,t))
 \Big]_{+} \\
 \leq
 \mathbb{E}[U_{\bm \theta}] \sum_{m=1}^{M} \mathbb{E} \Big[  \big(f_m({\bf x})  - \mu^t_m({\bf x}) - \sqrt{\beta_t}\sigma_m^t({\bf x}) \big).(1-C({\bf x},t))  \Big]_{+}
\end{align*}
Since $C({\bf x},t)$ is bounded as $C({\bf x},t) \in (0,1)$:
\begin{align*}
 \mathbb{E}[U_{\bm \theta}] \sum_{m=1}^{M} \mathbb{E} \Big[ \big(f_m({\bf x})  - \mu^t_m({\bf x}) - \sqrt{\beta_t}\sigma_m^t({\bf x}) \big).(1-C({\bf x},t)) \Big]_{+} \\
\leq  \mathbb{E}[U_{\bm \theta}] \sum_{m=1}^{M} \mathbb{E} \Big[  f_m({\bf x})  - \mu^t_m({\bf x}) - \sqrt{\beta_t}\sigma_m^t({\bf x}) \Big]_{+}
\end{align*}
Following the Gaussian process, we know that $\big[\big(f_m({\bf x})  - \mu^t_m({\bf x}) - \sqrt{\beta_t}\sigma_m^t({\bf x})\big)_+| \mathcal{H}_t\big] \sim \mathcal{N}\Big(- \sqrt{\beta_t}\sigma_m^t({\bf x}),\big(\sigma_m^t({\bf x})\big)^2\Big)$. As it has been described in \cite{Kand2018}:
\[
\big[\big(f_m({\bf x})  - \mu^t_m({\bf x}) - \sqrt{\beta_t}\sigma_m^t({\bf x})\big)_+| \mathcal{H}_t\big]\\
\leq \frac{\sigma_m^t({\bf x})}{\sqrt{2\pi}} e^{-\frac{\beta_t}{2}} \leq
\frac{1}{t^2 |\mathcal{X}|}
\]
Using tower property of expectation:
\[
\mathbb{E} \Big[
\big(  
S_{{\bm \theta}_t} (f({\bf x}^*_t)) - \alpha({\bf x}^*_t,{\bm \theta}_t,t)\big).(1-C({\bf x}^*_t,t))
 \Big] \leq 
 \mathbb{E}[U_{\bm \theta}]\frac{M}{t^2}
\]
As $\sum_{t=1}^{T} \frac{1}{t^2} \leq \frac{\pi^2}{6}$, finally:
\begin{align*}
\mathbb{E} \Bigg[
\sum_{t=1}^{T} \big(  
S_{{\bm \theta}_t} (f({\bf x}^*_t)) - \alpha({\bf x}^*_t,{\bm \theta}_t,t)\big).(1-C({\bf x}^*_t,t))
 \Bigg] \\
 \leq \frac{\mathbb{E}[U_{\bm \theta}]\times \pi^2}{6}M 
\end{align*}
\end{proof}
\begin{cor_sup}
It can be proved that $\mathrm{(I)}$ is bounded as:
\begin{align*}
\mathbb{E} \Bigg[
\sum_{t=1}^{T} \Big(  
\alpha({\bf x}_t,{\bm \theta}_t,t) - S_{{\bm \theta}_t} (f({\bf x}_t))\Big).\Big(1-C({\bf x}_t,t)\Big)
 \Bigg] \\ 
 \leq {\bar{U}}_{\bm \theta}\Big(MT\beta_T \sum_{m=1}^{M} \frac{\gamma_{T(m)}}{ln(1+\sigma^{-2}_{m})}\Big)^{\frac{1}{2}} + \frac{\pi^2}{6}M \mathbb{E}[{U}_{\bm \theta}] 
\end{align*}
where 
${\bar{U}}_{\bm \theta} = \mathbb{E} \big[ \sqrt{\frac{1}{T} \sum_{t=1}^{T} U^2_{{\bm \theta}_t}} \big]$.
\label{cor:Sup2}
\end{cor_sup}

\begin{proof}
We are starting the proof by:
\begin{align*}
\mathbb{E} \Bigg[
\sum_{t=1}^{T} \Big(  
\alpha({\bf x}_t,{\bm \theta}_t,t) - S_{{\bm \theta}_t} (f({\bf x}_t))\Big).\Big(1-C({\bf x}_t,t)\Big)
  | \mathcal{H}_t \Bigg] \\ 
 \leq \mathbb{E} \Bigg[
U_{{\bm \theta}_t} \sum_{m=1}^{M} \Big(
\mu_m^t ({\bf x}_t) + \sqrt{\beta_t} \sigma_m^t({\bf x}_t) - f_m({\bf x}_t)
\Big).\Big(1-C({\bf x}_t,t)\Big) + \\
U_{{\bm \theta}_t} \sum_{m=1}^{M} \Big(
f_m({\bf x}_t) - \mu_m^t ({\bf x}_t) - \sqrt{\beta_t} \sigma_m^t({\bf x}_t) 
\Big)_+.\Big(1-C({\bf x}_t,t)\Big)
 \Bigg]\\
\end{align*}
Based on the corollary \ref{cor:Sup1}, $\big(f_m({\bf x})  - \mu^t_m({\bf x}) - \sqrt{\beta_t}\sigma_m^t({\bf x})\big)_+ \leq  \frac{1}{t^2 |\mathcal{X}|} $ and since we know the cost function is bounded $C({\bf x}_t,t) \in (0,1)$, the inequality can be written as:
\begin{align*}
 \leq \mathbb{E} \Bigg[ 
 U_{{\bm \theta}_t} \sum_{m=1}^{M} \sqrt{\beta_t} \sigma_m^t ({\bf x}_t) 
 \Bigg] + \mathbb{E} \Bigg[  
 \frac{M U_{{\bm \theta}_t}}{t^2 |\mathcal{X}|}
 \Bigg] 
\end{align*}
Adding the sum over $T$ iterations:
\begin{align*}
\mathbb{E} \Bigg[
\sum_{t=1}^{T} \Big(  
\alpha({\bf x}_t,{\bm \theta}_t,t) - S_{{\bm \theta}_t} (f({\bf x}_t))\Big).\Big(1-C({\bf x}_t,t)\Big)
 \Bigg] \\
 \leq
 \mathbb{E} \Bigg[
 \sum_{m=1}^{M} \sum_{t=1}^{T} U_{{\bm \theta}_t} \sqrt{\beta_t} \sigma_m^t ({\bf x}_t)\Bigg] + \mathbb{E} \Bigg[\sum_{t=1}^{T} \frac{MU_{{\bm \theta}_t}}{t^2 |\mathcal{X}|} 
 \Bigg]
\end{align*}
\begin{align*}
 \leq
 \mathbb{E} \Bigg[
 \sum_{m=1}^{M} \sum_{t=1}^{T} U_{{\bm \theta}_t} \sqrt{\beta_t} \sigma_m^t ({\bf x}_t)\Bigg] + 
 \frac{\pi^2}{6}\frac{M\mathbb{E}[U_{{\bm \theta}_t}]}{|\mathcal{X}|}
\end{align*}
Based on the sums over $T$ and $M$, \cite{Kand2018} and \cite{srinivas2009gaussian} proved that:
\begin{align*}
 \mathbb{E} \Bigg[
 \sum_{m=1}^{M} \sum_{t=1}^{T} U_{{\bm \theta}_t} \sqrt{\beta_t} \sigma_m^t ({\bf x}_t)\Bigg] \\
 \leq
 \mathbb{E} \Bigg[
 \Big(
 M\beta_T\sum_{t=1}^{T} U^2_{{\bm \theta}_t}
 \Big)^\frac{1}{2}
 \Big(
 \sum_{m=1}^{M} \frac{\gamma_{T(m)}}{ln(1+\sigma_m^{-2})}
 \Big)^\frac{1}{2}
 \Bigg]
\end{align*}
So it can be concluded that
\begin{align*}
\mathbb{E} \Bigg[
\sum_{t=1}^{T} \Big(  
\alpha({\bf x}_t,{\bm \theta}_t,t) - S_{{\bm \theta}_t} (f({\bf x}_t))\Big).\Big(1-C({\bf x}_t,t)\Big)
 \Bigg]\\
 \leq
\bar{U}_{\bm \theta}\Big(MT\beta_T \sum_{m=1}^{M} \frac{\gamma_{T(m)}}{ln(1+\sigma^{-2}_{m})}\Big)^{\frac{1}{2}} + \frac{\pi^2}{6}M \mathbb{E}[{U}_{\bm \theta}]
\end{align*}
Finally based on the corollary \ref{cor:Sup1} and corollary \ref{cor:Sup2}, it proves that:
\begin{align*}
\sum_{t=1}^{T} \Big(\mathrm{\maxx_{{\bf x} \in \mathbb{X}}}\ \  S_{{\bm \theta}_t} \big(f({\bf x})\big)\big(1- C({\bf x},t)\big)-S_{{\bm \theta}_t} \big(f({\bf x}_t)\big)\big(1-C({\bf x_t},t)\big)\Big)
\\ \leq
\bar{U}_{\bm \theta}\Big(MT\beta_T \sum_{m=1}^{M} \frac{\gamma_{T(m)}}{ln(1+\sigma^{-2}_{m})}\Big)^{\frac{1}{2}} + \frac{\pi^2}{3}M \mathbb{E}[{U}_{\bm \theta}]
\end{align*}
where ${\bar{U}}_{\bm \theta}$ is defined as ${\bar{U}}_{\bm \theta} = \mathbb{E} \big[ \sqrt{\frac{1}{T} \sum_{t=1}^{T} U^2_{{\bm \theta}_t}} \big]$.
\end{proof}

\subsection{Additional Experiments}
Figure \ref{fig:exp2} shows the results for minimising Matyas function as first objective and Booth function \cite{jamil2013literature} as the second objective:
\[
f_1(\mathbf x) = 0.26(x_1^2+x_2^2) - 0.48x_1x_2
\]
\[
f_2(\mathbf x) = (x_1+2x_2-7)^2 - (2x_1+x_2-5)^2  
\]
Figure \ref{fig:2x1} show the whole Pareto front solution for this problem. A comparison between Figure \ref{fig:2x5} and Figure \ref{fig:2x7} shows that CA-MOBO finds the regions of Pareto front in a different regions of objective space compared to MOBO that found the solutions with higher exploration in expensive regions of search space due to lack of cost-aware constraints. 
Comparing Figure \ref{fig:2x6} and \ref{fig:2x8}, $\sum_{t=1}^{T=500}\ x_1 < \sum_{t=1}^{T=500}\ x_2,\ \forall t$ for CA-MOBO which indicates that the expensive dimension of search space has been used less than  the cheaper dimension of search space, however in Figure \ref{fig:2x8}, $\sum_{t=1}^{T=500}\ x_2 < \sum_{t=1}^{T=500}\ x_1,\ \forall t$. As Figure \ref{fig:2x2} demonstrates, even by imposing cost-aware constraints, CA-MOBO achieved better dominance in hypervolume comparing to MO-UCB. Figure \ref{fig:2x3} and Figure \ref{fig:2x4} illustrate the average and cumulative regret respectively.
\end{document}